\documentclass[10pt,twocolumn,letterpaper]{article}

\usepackage{cvpr}
\usepackage{times}
\usepackage{epsfig}
\usepackage{graphicx}
\usepackage{amsmath}
\usepackage{amssymb}

\usepackage{xcolor}
\usepackage{subfigure}
\usepackage{multirow}
\usepackage{booktabs}
\usepackage{verbatim}
\usepackage{siunitx}
\usepackage{textcomp}
\usepackage[title]{appendix}

\newcommand{\vv}[1]{\boldsymbol{#1}}

\usepackage[pagebackref=true,breaklinks=true,letterpaper=true,colorlinks,bookmarks=false]{hyperref}

\cvprfinalcopy 

\def\cvprPaperID{6983} 

\ifcvprfinal\fi
\begin{document}

\title{MaskFlownet: Asymmetric Feature Matching with Learnable Occlusion Mask}

\author{Shengyu Zhao\thanks{Equal contribution. \textsuperscript{$\dagger$}Corresponding author (xuyan04@gmail.com).
\newline Shengyu Zhao, Yilun Sheng, and Yue Dong are with Institute for Interdisciplinary Information Sciences, Tsinghua University, Beijing 100084, China.
Shengyu Zhao, Yilun Sheng, Eric I-Chao Chang, and Yan Xu are with Microsoft Research, Beijing 100080, China.
Yan Xu is with State Key Laboratory of Software Development Environment and Key Laboratory of Biomechanics and Mechanobiology of Ministry of Education and Research Institute of Beihang University in Shenzhen and Beijing Advanced Innovation Center for Biomedical Engineering, Beihang University, Beijing 100191, China.
\newline  This work is supported by the National Science and Technology Major Project of the Ministry of Science and Technology in China under Grant 2017YFC0110903,  Microsoft Research under the eHealth program, the National Natural Science Foundation in China under Grant 81771910, the Fundamental Research Funds for the Central Universities of China under Grant SKLSDE-2017ZX-08 from the State Key Laboratory of Software Development Environment in Beihang University in China, the 111 Project in China under Grant B13003.}
\quad Yilun Sheng\footnotemark[1] \quad Yue Dong \quad Eric I-Chao Chang \quad Yan Xu\footnotemark[2]
}

\maketitle
\thispagestyle{empty}

\begin{abstract}
Feature warping is a core technique in optical flow estimation; however, the ambiguity caused by occluded areas during warping is a major problem that remains unsolved. In this paper, we propose an asymmetric occlusion-aware feature matching module, which can learn a rough occlusion mask that filters useless (occluded) areas immediately after feature warping without any explicit supervision. The proposed module can be easily integrated into end-to-end network architectures and enjoys performance gains while introducing negligible computational cost. The learned occlusion mask can be further fed into a subsequent network cascade with dual feature pyramids with which we achieve state-of-the-art performance. At the time of submission, our method, called MaskFlownet, surpasses all published optical flow methods on the MPI Sintel, KITTI 2012 and 2015 benchmarks. Code is available at \url{https://github.com/microsoft/MaskFlownet}.
\end{abstract}


\section{Introduction}


\begin{figure}[t]
\begin{center}
\subfigure[Image warping and \textbf{masking}.]{ \label{fig:image_masking}
  \includegraphics[width=1.0\linewidth]{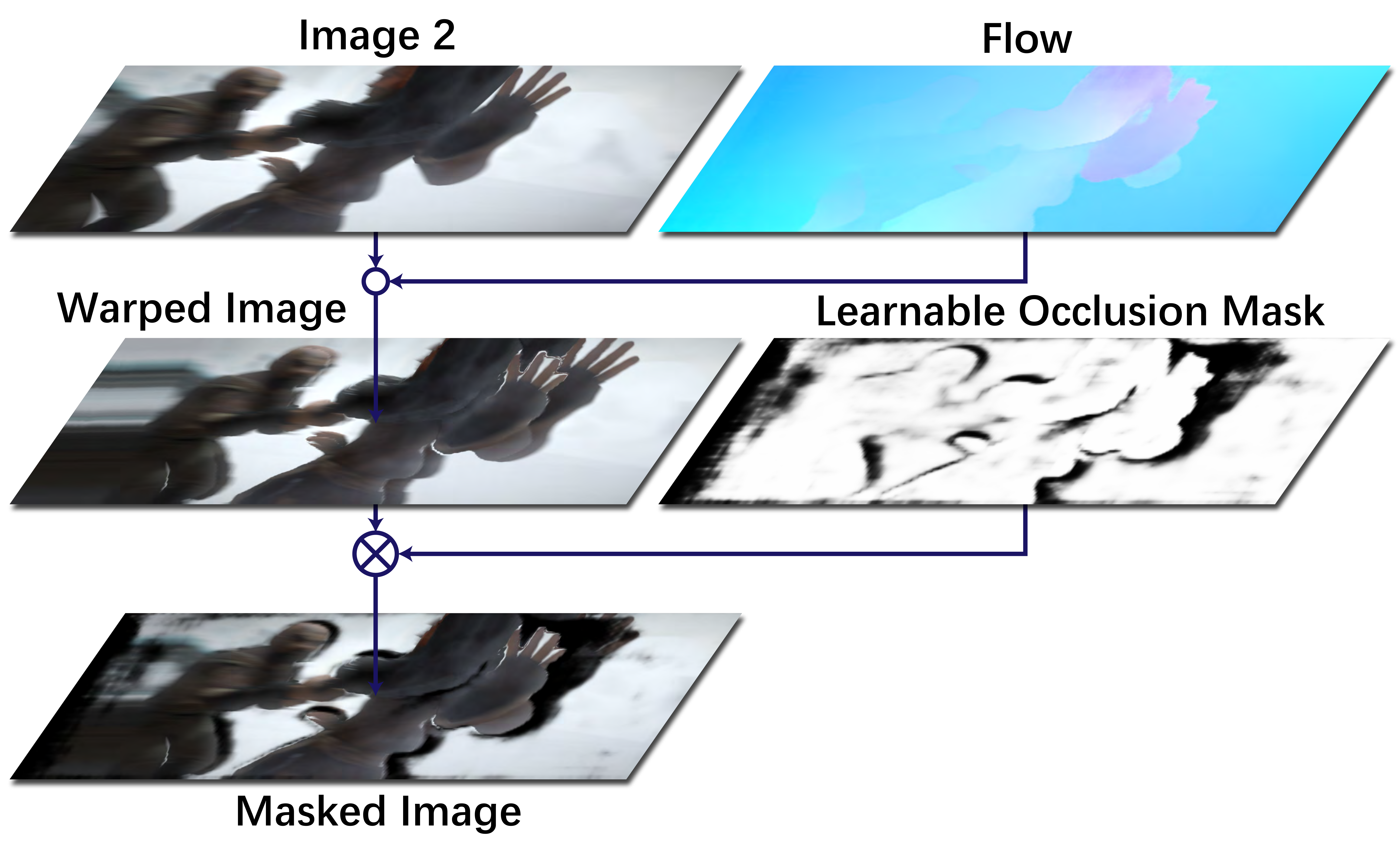}}
\subfigure[Feature warping and \textbf{masking}.]{ \label{fig:feature_masking}
  \includegraphics[width=1.0\linewidth]{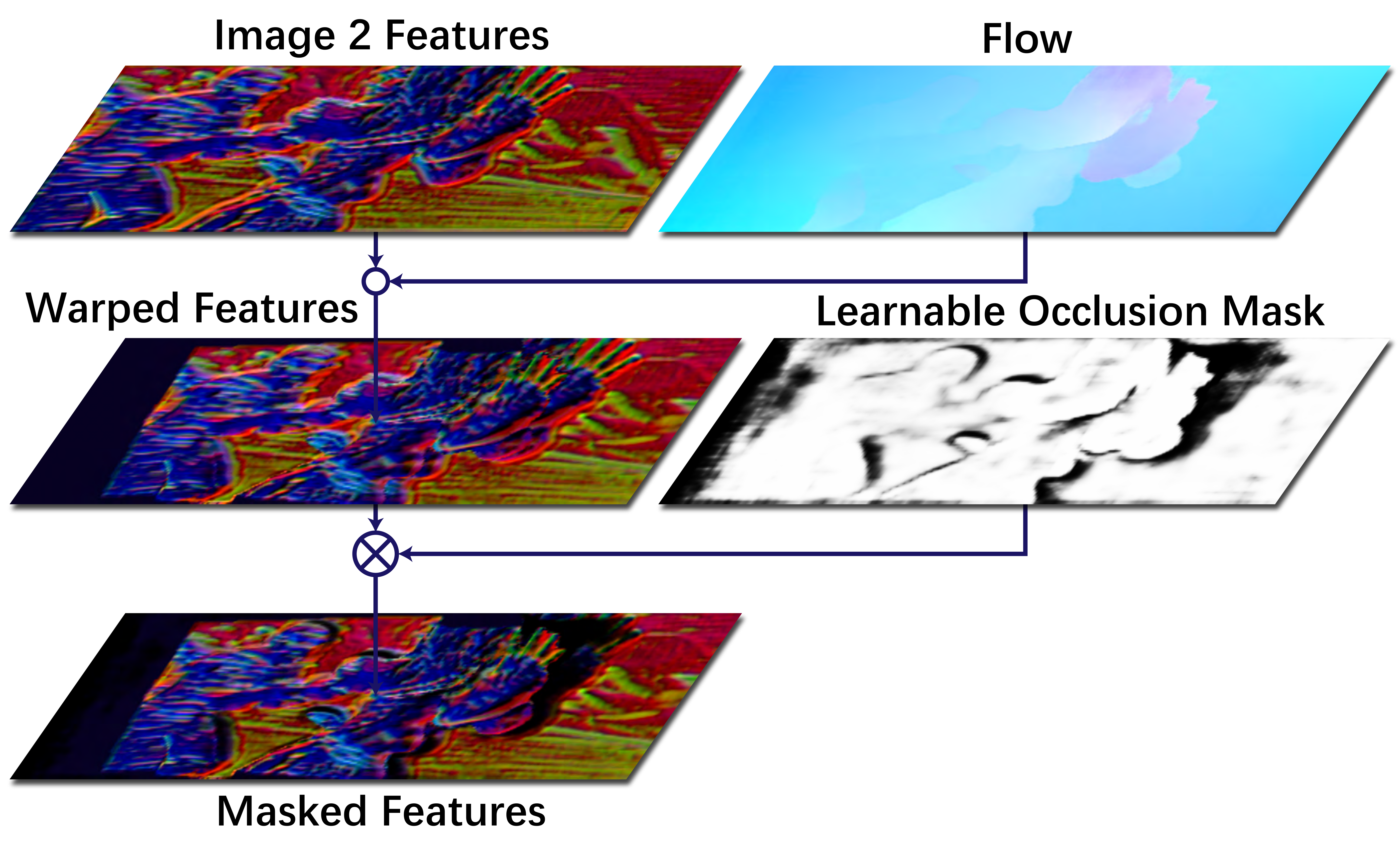}}
\end{center}
  \caption{\textbf{Motivation of the learnable occlusion mask.} (a) Image warping induces ambiguity in the occluded areas (see the doubled hands). (b) The same problem exists in the feature warping process. Such areas can be masked without any explicit supervision.}
\label{fig:masking}
\end{figure}

\begin{figure*}[t]
\begin{center}
   \includegraphics[width=1.0\linewidth]{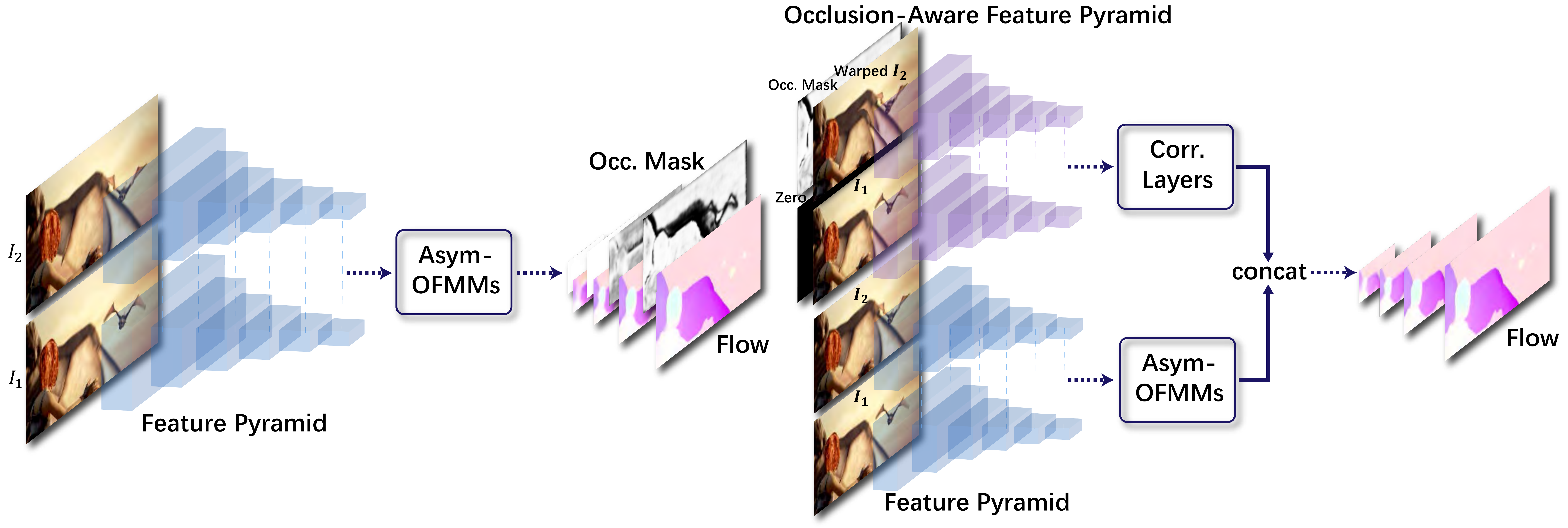}
\end{center}
   \caption{\textbf{The overall architecture of MaskFlownet.} MaskFlownet consists of two stages --- the first end-to-end network named MaskFlownet-S (left), and the second cascaded network (right) that aims to perform refinements using dual pyramids. Dashed lines across pyramids represent shared weights. MaskFlownet generally utilizes the proposed AsymOFMM whenever possible. The learnable occlusion mask is coarse-to-fine predicted and fed into the new occlusion-aware feature pyramid. See \S\ref{sec:network} for the network details. }
\label{fig:overall_architecture}
\end{figure*}


Optical flow estimation is a core problem in computer vision and a fundamental building block in many real-world applications~\cite{Bonneel2015Video,Menze2015Driving,Simonyan2014Action}.
Recent development towards fast, accurate optical flow estimation has witnessed great progress of learning-based methods using a principled network design --- feature pyramid, warping, and cost volume --- proposed by PWC-Net~\cite{Sun2018PWC} and LiteFlowNet~\cite{Hui2018Liteflownet}, and used in many follow-up works~\cite{Hui2019Liteflownet2,Hur2019IRR,Liu2019SelFlow,Neoral2018Continual,Sun2018PWC+}.
Feature warping effectively resolves the long-range matching problem between the extracted feature maps for the subsequent cost volume computation.
However, we observe that a major problem of the warping operation is that it introduces unreliable information in the presence of occlusions. As shown in Fig.~\ref{fig:masking}, the warped image as well as the warped feature map can be even ``doubled'' at the occluded areas (also called the \emph{ghosting effect}). It remains unclear that whether the source image would be mismatched to such areas yet raises a natural question: \emph{are they really distinguishable without being supervised of occlusions?}

We answer this question positively by showing that the network can indeed learn to \emph{mask} such areas without any explicit supervision. Rather than enforcing the network to distinguish useful parts from those confusing information, we propose to apply a multiplicative learnable occlusion mask immediately on the warped features (see Fig.~\ref{fig:masking}). We can see that there is a clear distinction between the black and white areas in the learned occlusion mask, indicating that there exists solid gradient propagation. The masked image (features) has much cleaner semantics, which could potentially facilitate the subsequent cost volume processing.

The masking process interprets how those areas can be distinguished in a clear way. While previous works commonly believe that the feature symmetricity is crucial for the cost volume processing, we in contrast demonstrate that the network further benefits from a simple \emph{asymmetric} design despite the explicit masking. The combined \emph{asymmetric occlusion-aware feature matching module} (AsymOFMM) can be easily integrated into end-to-end network architectures and achieves significant performance gains.

We demonstrate how the proposed AsymOFMM would contribute to the overall performance using a two-stage architecture named \emph{MaskFlownet} (see Fig.~\ref{fig:overall_architecture}). MaskFlownet is trained on standard optical flow datasets (not using the occlusion ground truth), and predicts the optical flow together with a rough occlusion mask in a single forward pass. At the time of submission, MaskFlownet surpasses all published optical flow methods on the MPI Sintel (on both clean and final pass), KITTI 2012 and 2015 benchmarks while using only \emph{two-frame} inputs with no additional assumption.

\section{Related Work}

\paragraph{Optical Flow Estimation.}

Conventional approaches formulate optical flow estimation as an energy minimization problem based on brightness constancy and spatial smoothness since~\cite{Horn1981Flow} with many follow-up improvements~\cite{Brox2004High,Memin1998Dense,Wedel2009Structure}. Estimating optical flow in a coarse-to-fine manner achieves better performance since it better solves large displacements~\cite{Brox2010DescripMatch,Weinzaepfel2013DeepFlow}. Later works propose to use CNN extractors for feature matching~\cite{Bailer2017CNNMatch,Xu2017DirectCost}. However, their high accuracy is at the cost of huge computation, rendering those kinds of methods impractical in real-time settings.

An important breakthrough of deep learning techniques in optical flow estimation is made by FlowNet~\cite{Dosovitskiy2015Flownet}, which proposes to train end-to-end CNNs on a synthetic dataset and first achieves a promising performance. Although they only investigate two types of simple CNNs (FlowNetS and FlowNetC), the correlation layer in FlowNetC turns out to be a key component in the modern architectures. Flownet2~\cite{Ilg2017Flownet2} explores a better training schedule and makes significant improvements by stacking multiple CNNs which are stage-wise trained after fixing the previous ones.

SpyNet~\cite{Ranjan2017SpyNet} explores a light-weight network architecture using feature pyramid and warping, but the learned features are not correlated so it can only achieve a comparable performance to FlowNetS. PWC-Net~\cite{Sun2018PWC} and LiteFlowNet~\cite{Hui2018Liteflownet} present a compact design using feature pyramid, warping, and cost volume, and achieve remarkable performance over most conventional methods while preserving high efficiency, and they further make some slight improvements in the later versions~\cite{Hui2019Liteflownet2,Sun2018PWC+}. VCN~\cite{Yang2019VCN} recently exploits the high-dimensional invariance during cost volume processing and achieves state-of-the-art performance. Note that this paper focuses on the feature matching process prior to the correlation layer, which is independent of the improvement made by VCN. To our knowledge, none of those works realizes that an asymmetric design of the feature matching process can achieve better performance.

\paragraph{Occlusions and Optical Flow.}

Occlusions and optical flow are closely related. Optical flow methods such as FlowNet and FlowNet2 can be easily extended to joint optical flow and occlusion estimation with slight modifications as proposed in~\cite{Ilg2018DispNet2,Mayer2016Things3D}. IRR-PWC~\cite{Hur2019IRR} presents an iterative residual refinement approach with joint occlusion estimation using bilateral refinement. However, all those methods can only explicitly learn from the ground-truth occlusions, which require additional efforts on the training labels that limit their applicability~\cite{Hur2019IRR,Mayer2016Things3D}.

Unsupervised or self-supervised learning of optical flow is another promising direction. Handling occlusions is clearly a vital aspect in such setting, since the brightness error does not make sense at occluded pixels. Some initial works show the feasibility of the unsupervised learning guided by the photometric loss~\cite{Jason2016UnBrightness,Ren2017UnDeep}. Later works realize that the occluded pixels should be excluded from the loss computation~\cite{Meister2018UnFlow,Wang2018Occ}. Occlusions also facilitate multi-frame estimation~\cite{Janai2018UnMulti,Liu2019SelFlow,Neoral2018Continual}. However, the only occlusion estimation approach used by those kinds of methods is the forward-backward consistency~\cite{Sundaram2010Dense} that requires bidirectional flow estimation, which limits its flexibility and could lead to noisy predictions. We would like to remark that our promising approach can jointly estimate occlusions without any explicit supervision in a single forward pass, which we expect can be helpful to future unsupervised or self-supervised learning methods.

\paragraph{Occlusion-Aware Techniques in Other Applications.}

Occlusions commonly exist in object detection and might affect the performance of standard approaches in some scenarios, e.g., crowded pedestrian detection~\cite{Zhang2018OccRCNN}. Recent works propose to explicitly learn a spatial attention mask that highlights the foreground area for occluded pedestrian detection~\cite{Pang2019PedMaskAtt,Zhang2018PedGuidedAtt}, which requires additional supervising information. Occlusions in face recognition is also a major problem which can be addressed by the guided mask learning~\cite{Song2019FaceOcc}. Our work is also related to the attention mechanism in computer vison~\cite{Wang2018NonLocal}, which addresses a different problem of capturing \emph{pixel-wise} long-range dependencies. None of those works realizes a global attention mask can be learned to filter occluded areas with no explicit supervision.

\section{Occlusion-Aware Feature Matching}

\begin{figure}[t]
\begin{center}
   \includegraphics[width=1.0\linewidth]{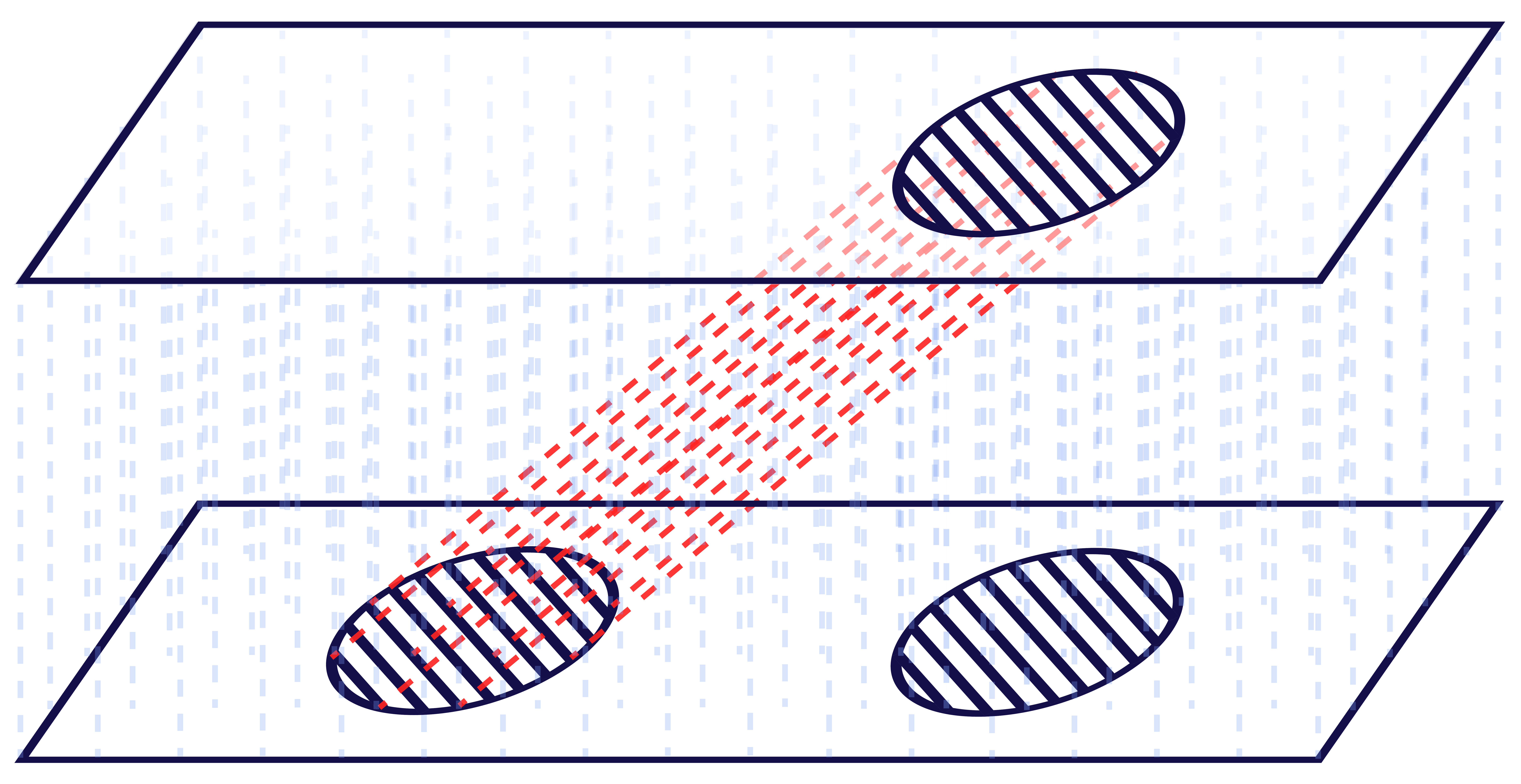}
\end{center}
   \caption{\textbf{A simplified case of occlusions.} The top image is warped to the bottom image according to the illustrated flow. The foreground object (shaded area) generates a large displacement (tracked by the red lines) while the background stays still (tracked by the blue lines). However, a copy of the foreground object still stays at the occluded area after warping.}
\label{fig:occ_example}
\end{figure}

Given an image pair $I_1$ (the source image) and $I_2$ (the target image), the task is to estimate the flow displacement $\vv \phi$, representing the correspondence between $I_1(\vv x)$ and $I_2(\vv x + \vv \phi(\vv x))$. Image warping is the process of constructing
\begin{align}
(\vv \phi \circ I_2)(\vv x) \triangleq I_2(\vv x + \vv \phi(\vv x)) \label{eq:warp}
\end{align}
using the estimated displacement $\vv \phi$, and ideally we have $I_1(\vv x) \approx (\vv \phi \circ I_2)(\vv x)$ at all non-occluded pixels $\vv x$. This operation is differentiable w.r.t.\@ both inputs because non-integral points can be dealt with bilinear interpolation.
Feature warping is similarly defined by replacing $I_2$ in Eq.\@ (\ref{eq:warp}) with the extracted feature map.

\begin{figure}[t]
\begin{center}
\subfigure[Feature matching module (FMM) as used in PWC-Net~\cite{Sun2018PWC}.]{ \label{fig:FMM}
   \includegraphics[width=1.0\linewidth]{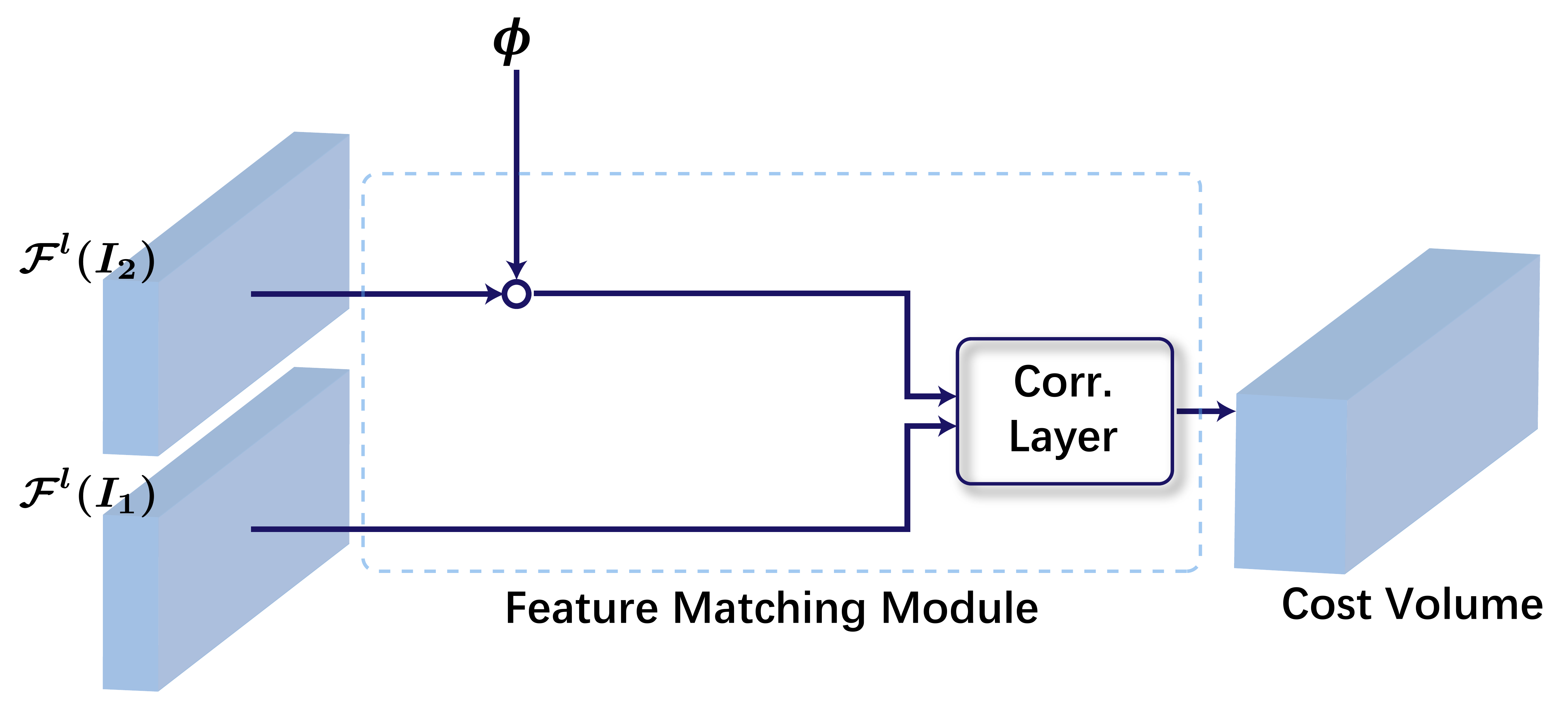}}
\subfigure[Occlusion-aware feature matching module (OFMM).]{ \label{fig:OFMM}
   \includegraphics[width=1.0\linewidth]{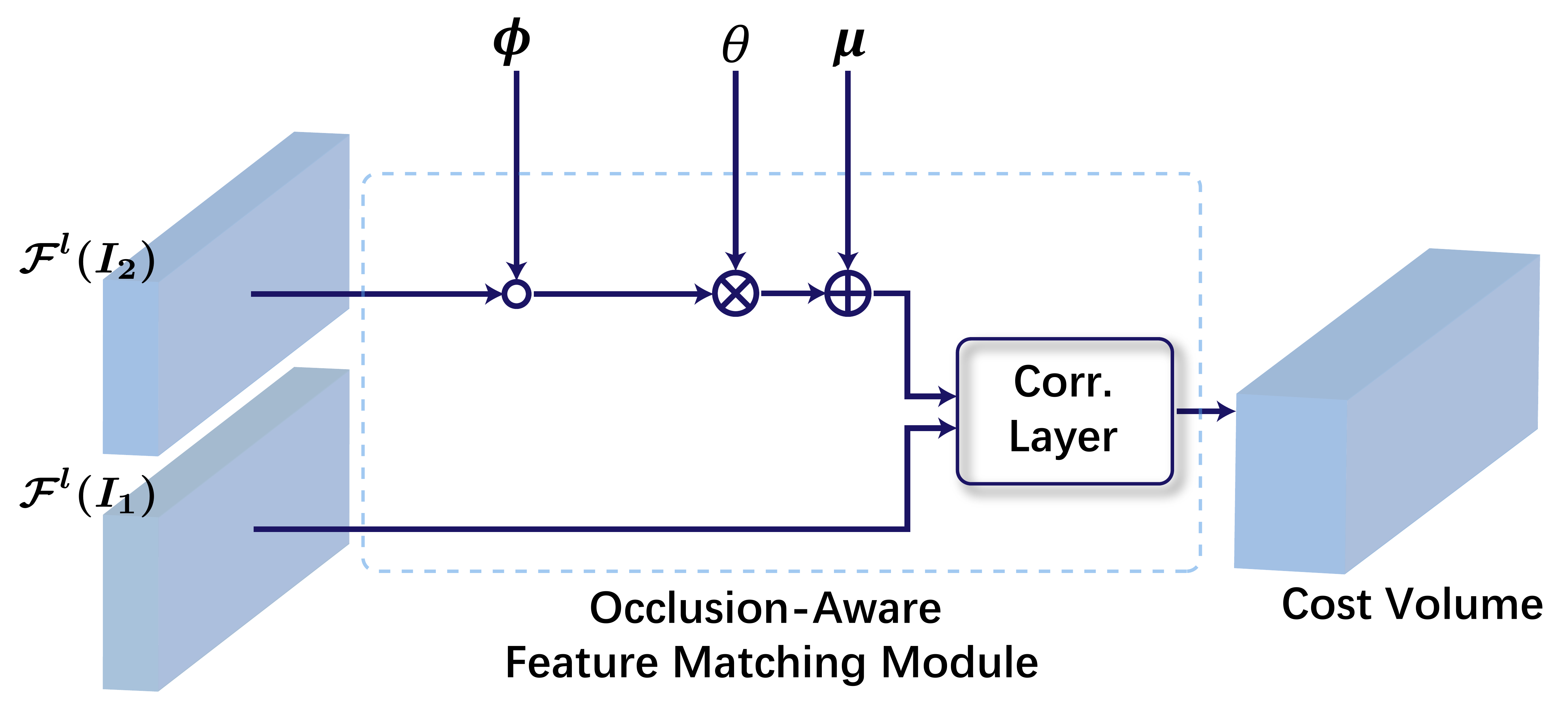}}
\subfigure[Asymmetric occlusion-aware feature matching module (AsymOFMM).]{ \label{fig:AsymOFMM}
   \includegraphics[width=1.0\linewidth]{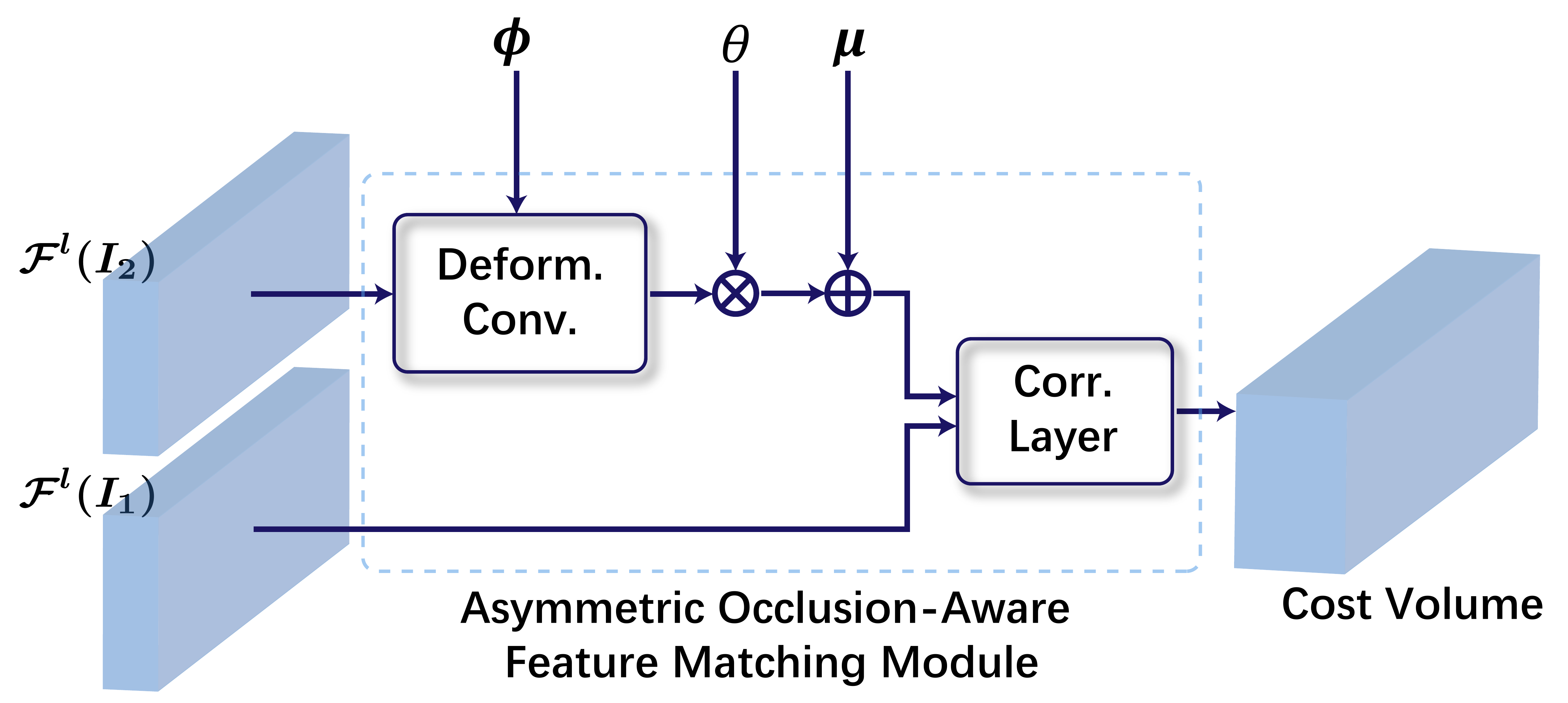}}
\end{center}
   \caption{\textbf{Feature matching modules.} The figures illustrate the proposed AsymOFMM and OFMM in comparison with the FMM. FMM warps the target feature maps with the flow displacement $\vv \phi$. OFMM introduces the multiplicative learnable occlusion mask $\theta$ followed by the additive trade-off term $\vv \mu$. AsymOFMM further replaces the warping operation by a deformable convolution.}
\end{figure}

Feature warping followed by a correlation layer is the common practice to compute the cost volume for non-local feature matching in recent works~\cite{Hui2018Liteflownet,Hui2019Liteflownet2,Hur2019IRR,Liu2019SelFlow,Neoral2018Continual,Sun2018PWC+,Sun2018PWC}. The feature extractor of an image is a pyramid of convolutional layers, which are proposed to be \emph{symmetric} for $I_1$ and $I_2$ in the sense that they share the \emph{same} convolution kernels. The cost volume at pyramid level $l$ can be formulated as
\begin{align}
\vv c(\mathcal{F}^l(I_1), \vv \phi \circ \mathcal{F}^l(I_2)),
\end{align}
where $\mathcal{F}^l$ denotes the shared feature extractor for level $l$, and $\vv \phi$ denotes the flow displacement predicted by the previous level. $\vv c$ represents the correlation layer that computes the element-wise dot product between the two feature maps within a maximum displacement. Fig.~\ref{fig:FMM} illustrates this process, which we call a feature matching module (FMM).

We observe that a major consequence caused by the warping operation is the ambiguity in the presence of occlusions. Fig.~\ref{fig:occ_example} illustrates a simplified example, where the foreground object has a large movement while the background stays still. During warping, a copy of the foreground object is revealed in the occluded background area. Such areas in the warped image (features) are useless and cause confusion to the subsequent flow inference.

\paragraph{Occlusion-Aware Feature Matching Module (OFMM).}

The occlusion-aware feature matching module incorporates a \emph{learnable occlusion mask} that filters useless information immediately after feature warping (see Fig.~\ref{fig:OFMM}). The warped feature tensor of shape $(B, C, H, W)$ is element-wise multiplied by the (soft) learnable occlusion mask $\theta$ of shape $(B, 1, H, W)$ with broadcasting and then added with an additional feature tensor $\vv \mu$ of shape $(B, C, H, W)$. The resulting cost volume at level $l$ is formulated as
\begin{align}
\vv c(\mathcal{F}^l(I_1), (\vv \phi \circ \mathcal{F}^l(I_2)) \otimes \theta \oplus \vv \mu).
\end{align}
$\theta$ is assumed to be within the range of $[0, 1]$. $\vv \mu$ acts as a trade-off term that facilitates the learning of occlusions, as it provides extra information at the masked areas.

OFMM learns to mask the occluded areas simply because it realizes they are useless comparing to the trade-off term, even if there is no explicit supervision to the occlusions at all. Although the OFMM itself might not contribute significantly to the performance, it can learn a rough occlusion mask at negligible cost, which can be further fed into the new occlusion-aware feature pyramid (see \S \ref{sec:network}).

\paragraph{Asymmetric Occlusion-Aware Feature Matching Module (AsymOFMM).}

We suggest that an asymmetric design of the feature extraction layers consistently gains the performance. Intuitively, the warping operation induces ambiguity to the occluded areas and breaks the symmetricity of the feature matching process, so an asymmetric design might be helpful to conquer this divergence.

Based on the OFMM, we introduce an extra convolutional layer prior to the warping operation, which is asymmetrically drawn on the feature extraction process of only $I_2$. In practice, we replace the extra convolutional layer and the warping operation by a \emph{deformable convolution}. In the general setting of deformable convolutional networks~\cite{Dai2017Deformable}, different locations in each of the convolution kernels are associated with different offsets, but here we introduce a specialized setting where each convolution kernel is warped in parallel according to the corresponding flow displacement at center. The deformable convolution slightly differs from the initial design since it reverts the order of convolution and (bilinear) interpolation, which is proved to be better in the experiments. As illustrated in Fig.~\ref{fig:AsymOFMM}, the resulting cost volume can be formulated as
\begin{align}
\vv c(\mathcal{F}^l(I_1), \mathcal{D}^l(\mathcal{F}^l(I_2), \vv \phi) \otimes \theta \oplus \vv \mu),
\end{align}
where $\mathcal{D}^l(\cdot, \vv \phi)$ denotes the deformable convolution layer at level $l$ using the displacement $\vv \phi$.


\section{MaskFlownet} \label{sec:network}

The overall architecture of the proposed MaskFlownet is illustrated in Fig.~\ref{fig:overall_architecture}. MaskFlownet consists of two cascaded subnetworks. The first stage, named MaskFlownet-S, generally inherits the network architecture from PWC-Net~\cite{Sun2018PWC}, but replaces the feature matching modules (FMMs) by the proposed AsymOFMMs.

\begin{figure}[t]
\begin{center}
   \includegraphics[width=1.0\linewidth]{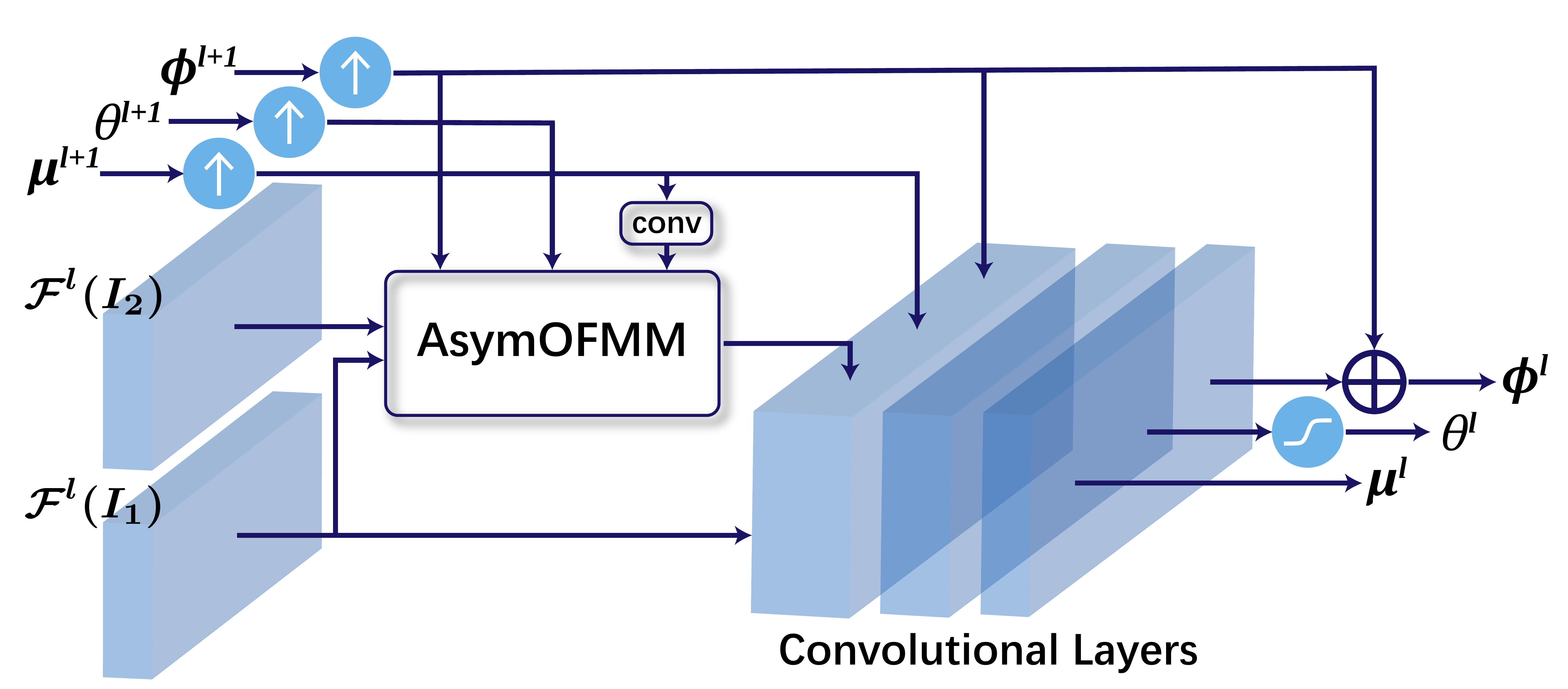}
\end{center}
   \caption{\textbf{Network connections at each level.} This figure adopts to the first stage (MaskFlownet-S). The learnable occlusion mask is generated through the sigmoid activation from the previous level, and then upsampled and fed into the AsymOFMM.}
\label{fig:network_details}
\end{figure}

MaskFlownet-S first generates a 6-level shared feature pyramid as PWC-Net, and then makes predictions from level 6 to 2 in a coarse-to-fine manner. The final predictions at level 2 are 4-time upsampled to level 0. Fig.~\ref{fig:network_details} illustrates the network details at each level $l$ (modifications needed at level 6 and level 2). The previous level is responsible for providing $\vv \phi^{l+1}$, $\theta^{l+1}$, and $\vv \mu^{l+1}$, which are then upsampled and fed into the AsymOFMM. $\vv \phi^{l+1}, \theta^{l+1}$ are upsampled using bilinear interpolation; $\vv \mu^{l+1}$ is upsampled using a deconvolutional layer (of 16 channels), followed by a convolutional layer to match the channels of the pyramid feature extractor. All convolutional layers have a kernel size of 3; all deconvolutional layers have a kernel size of 4. The feature matching module at level 6 is simply a correlation layer since there is no initial flow for warping. The maximum displacement of the correlation layers is kept to be 4. The resulting cost volume is concatenated with $\mathcal{F}^l(I_1)$, the upsampled displacement, and the upsampled features, and then passed through $5$ densely connected convolutional layers as PWC-Net. The final layer predicts the flow displacement $\vv \phi^l$ with residual link from the previous flow estimation, the occlusion mask $\theta^l$ after the sigmoid activation, and the features $\vv \mu^l$ passing to the next level. Level 2 only predicts the flow displacement, ended with the context network (as PWC-Net) that produces the final flow prediction.

\paragraph{Occlusion-Aware Feature Pyramid.}

The learned occlusion mask, concatenated with the warped image, is fed into the occlusion-aware feature pyramid for the subsequent flow refinement. The occlusion mask is subtracted by $0.5$ before concatenation; a zero mask is concatenated with $I_1$ for symmetricity. The occlusion-aware pyramid extracts features for the concatenated images (both with 4 channels) with shared convolutional layers as usual. We suggest that the occlusion mask facilitates the feature representation of the warped image, given the vast existence of occluded areas during warping.

\paragraph{Cascaded Flow Inference with Dual Pyramids.}

We propose to cascade the network by utilizing \emph{dual} feature pyramids. The occlusion-aware feature pyramid provides abundant information about the warped image from the previous flow estimation for refinement, but it cannot feedback to the new coarse-to-fine flow predictions. Hence, we suggest that the network can still gain complementary information from the original feature pyramid.

The network architecture of this stage is similar to the former stage except some modifications and the incorporation of the new occlusion-aware feature pyramid (see Fig.~\ref{fig:overall_architecture}). The original feature pyramid is directly placed into this stage using the same parameters. The maximum displacement of all correlation layers in this stage is set to 2 since we expect it to perform mainly refinements. Correlation layers are used as the feature matching modules for the occlusion-aware feature pyramid since there is no need for feature warping. At each level, the resulting cost volumes from dual feature pyramids are concatenated together with the other terms including an extra flow predicted from the previous stage at the current level. As suggested in FlowNet2~\cite{Ilg2017Flownet2}, we fix all the parameters in the first stage (MaskFlownet-S) when training the whole MaskFlownet.

\section{Experiments}

\subsection{Implementation}

We implement a Python-trainable code framework using MXNet~\cite{Chen2015MXNet}. Mostly, we follow the training configurations as suggested in PWC-Net+~\cite{Sun2018PWC+,Sun2018PWC} and FlowNet2~\cite{Ilg2017Flownet2}. More details can be found in the supplementary material.

\paragraph{Training Schedule.} MaskFlownet-S is first trained on FlyingChairs~\cite{Dosovitskiy2015Flownet} and then tuned on FlyingThings3D~\cite{Mayer2016Things3D} following the same schedule as PWC-Net~\cite{Sun2018PWC}. When fine-tuning on Sintel, we use the same batch configuration (2 from Sintel, 1 from KITTI 2015, and 1 from HD1K) and a longer training schedule (1000k iterations) referring to the cyclic learning rate proposed by PWC-Net+~\cite{Sun2018PWC+}. When training the whole MaskFlownet, we fix all the parameters in MaskFlownet-S as suggested in FlowNet2~\cite{Ilg2017Flownet2} and follow again the same schedule except that it is shorter on FlyingChairs (800k iterations). For submission to KITTI, we fine-tune our model on the combination of KITTI 2012 and 2015 datasets based on the tuned checkpoint on Sintel, while the input images are resized to $1280 \times 576$ (before augmentation and cropping) since the decreased aspect ratio better balances the vertical and horizontal displacement.

\paragraph{Data Augmentation.} We implement geometric and chromatic augmentations referring to the implementation of FlowNet~\cite{Dosovitskiy2015Flownet} and IRR-PWC~\cite{Hur2019IRR}. We suppress the degree of augmentations when fine-tuning on KITTI as suggested.
For sparse ground-truth flow in KITTI, the augmented flow is weighted averaged based on the interpolated valid mask.

\paragraph{Training Loss.} We follow the multi-scale end-point error (EPE) loss when training on FlyingChairs and FlyingThings3D, and its robust version on Sintel and KITTI, using the same parameters as suggested in PWC-Net+~\cite{Sun2018PWC+}. Weight decay is disabled since we find it of little help.


\begin{table*}[t]
\begin{center}
\begin{tabular}{lccccccccc}
\toprule
\multirow{3}{*}{Method} & {Time} & \multicolumn{2}{c}{Sintel \emph{clean}} & \multicolumn{2}{c}{Sintel \emph{final}} & \multicolumn{2}{c}{KITTI 2012} & \multicolumn{2}{c}{KITTI 2015} \\
\cmidrule(lr){2-2}
\cmidrule(lr){3-4}
\cmidrule(lr){5-6}
\cmidrule(lr){7-8}
\cmidrule(lr){9-10}
{} & (s) & AEPE & AEPE & AEPE & AEPE & AEPE & AEPE & Fl-all & Fl-all \\
{} & {} & \emph{train} & \emph{test} & \emph{train} & \emph{test} & \emph{train} & \emph{test} & \emph{train} & \emph{test} \\
\midrule
FlowNetS~\cite{Dosovitskiy2015Flownet} & 0.01 & 3.66 & 6.16 & 4.76 & 7.22 & 6.07 & 7.6 & - & - \\
FlowNetC~\cite{Dosovitskiy2015Flownet} & 0.05 & 3.57 & 6.08 & 5.25 & 7.88 & 7.31 & - & - & - \\
FlowNet2~\cite{Ilg2017Flownet2} & 0.12 & \textbf{2.02} & 3.96 & \textbf{3.14} & 6.02 & 4.09 & 1.8 & 28.20\% & 11.48\% \\
SpyNet~\cite{Ranjan2017SpyNet} & 0.16 & 4.12 & 6.64 & 5.57 & 8.36 & 9.12 & 4.1 & - & 35.07\% \\
MR-Flow~\cite{Wulff2017MRFlow} & 480 & - & 2.53 & - & 5.38 & - & - & - & 12.19\% \\
LiteFlowNet~\cite{Hui2018Liteflownet} & 0.09 & 2.48 & 4.54 & 4.04 & 5.38 & 4.00 & 1.6 & 28.50\% & 9.38\% \\
LiteFlowNet2~\cite{Hui2019Liteflownet2} & 0.04 & 2.24 & 3.45 & 3.78 & 4.90 & 3.42 & 1.4 & 25.88\% & 7.74\% \\
PWC-Net~\cite{Sun2018PWC} & 0.03 & 2.55 & 3.86 & 3.93 & 5.13 & 4.14 & 1.7 & 33.67\% & 9.60\% \\
PWC-Net+~\cite{Sun2018PWC+} & 0.03 & - & 3.45 & - & 4.60 & - & 1.5 & - & 7.90\% \\
SelFlow~\cite{Liu2019SelFlow} & 0.09 & - & 3.74 & - & 4.26 & - & 1.5 & - & 8.42\% \\
VCN~\cite{Yang2019VCN} & 0.18 & 2.21 & 2.81 & 3.62 & 4.40 & - & - & 25.1\% & 6.30\% \\
MaskFlownet-S & 0.03 & 2.33 & 2.77 & 3.72 & 4.38 & 3.21 & \textbf{1.1} & 23.58\% & 6.81\% \\
MaskFlownet & 0.06 & 2.25 & \textbf{2.52} & 3.61 & \textbf{4.17} & \textbf{2.94} & \textbf{1.1} & \textbf{23.14\%} & \textbf{6.11\%} \\
\bottomrule
\end{tabular}
\end{center}
\caption{\textbf{Results of different methods on the MPI Sintel, KITTI 2012 and 2015 benchmarks.} Values listed in the \emph{train} columns only consider those models which are \emph{not} trained on the corresponding training set and thus comparable. AEPE: average end-point error over all valid pixels. Fl-all: percentage of optical flow outliers over all valid pixels. Running times are referred to~\cite{Sun2018PWC}; our time is measured on an NVIDIA TITAN Xp GPU, which is comparable to the NVIDIA TITAN X used by~\cite{Sun2018PWC}.}
\label{tab:main}
\end{table*}

\begin{figure*}[t]
\begin{center}
   \includegraphics[width=1.0\linewidth]{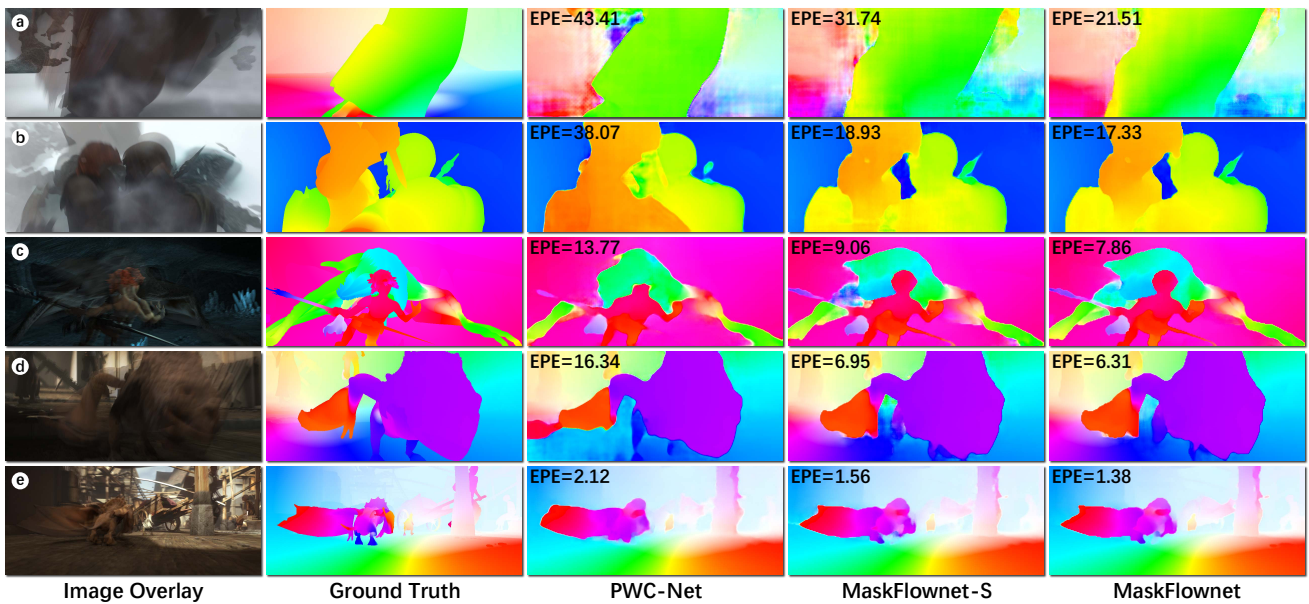}
\end{center}
   \caption{\textbf{Qualitative comparison among PWC-Net~\cite{Sun2018PWC}, MaskFlownet-S, and MaskFlownet.} Highlights for each row: (a) weakening the checkerboard effect; (b) separating background from two fighting figures; (c) preserving the weakly connected head of the moving figure; (d) maintaining a fluent flow for the background at left-bottom; (e) preserving boundary details of the flying creature and buildings.}
\label{fig:flow_visualization}
\end{figure*}

\begin{figure*}[t]
\begin{center}
   \includegraphics[width=1.0\linewidth]{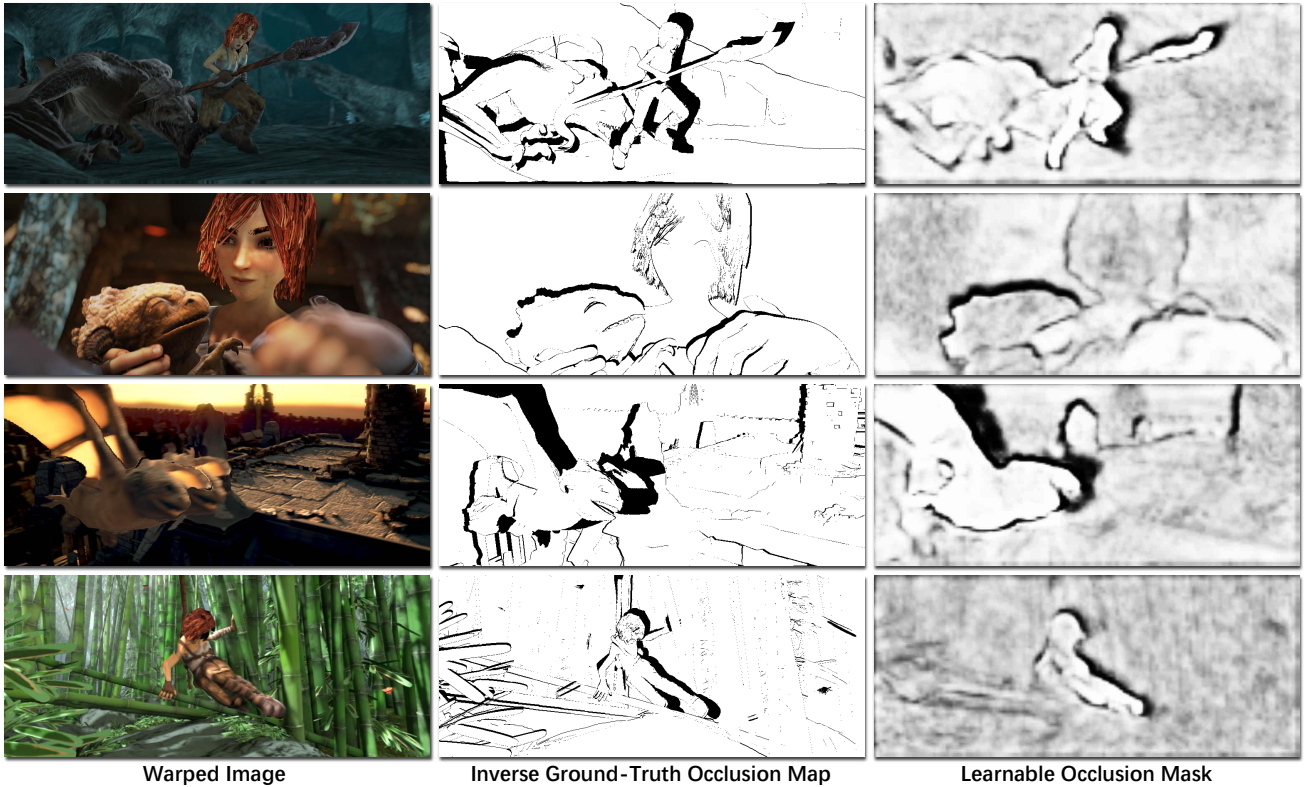}
\end{center}
   \caption{\textbf{Learnable occlusion mask.} MaskFlownet can jointly learn a rough occlusion mask without any explicit supervision.}
\label{fig:mask_visualization}
\end{figure*}

\begin{figure}[t]
\begin{center}
   \includegraphics[width=1.0\linewidth]{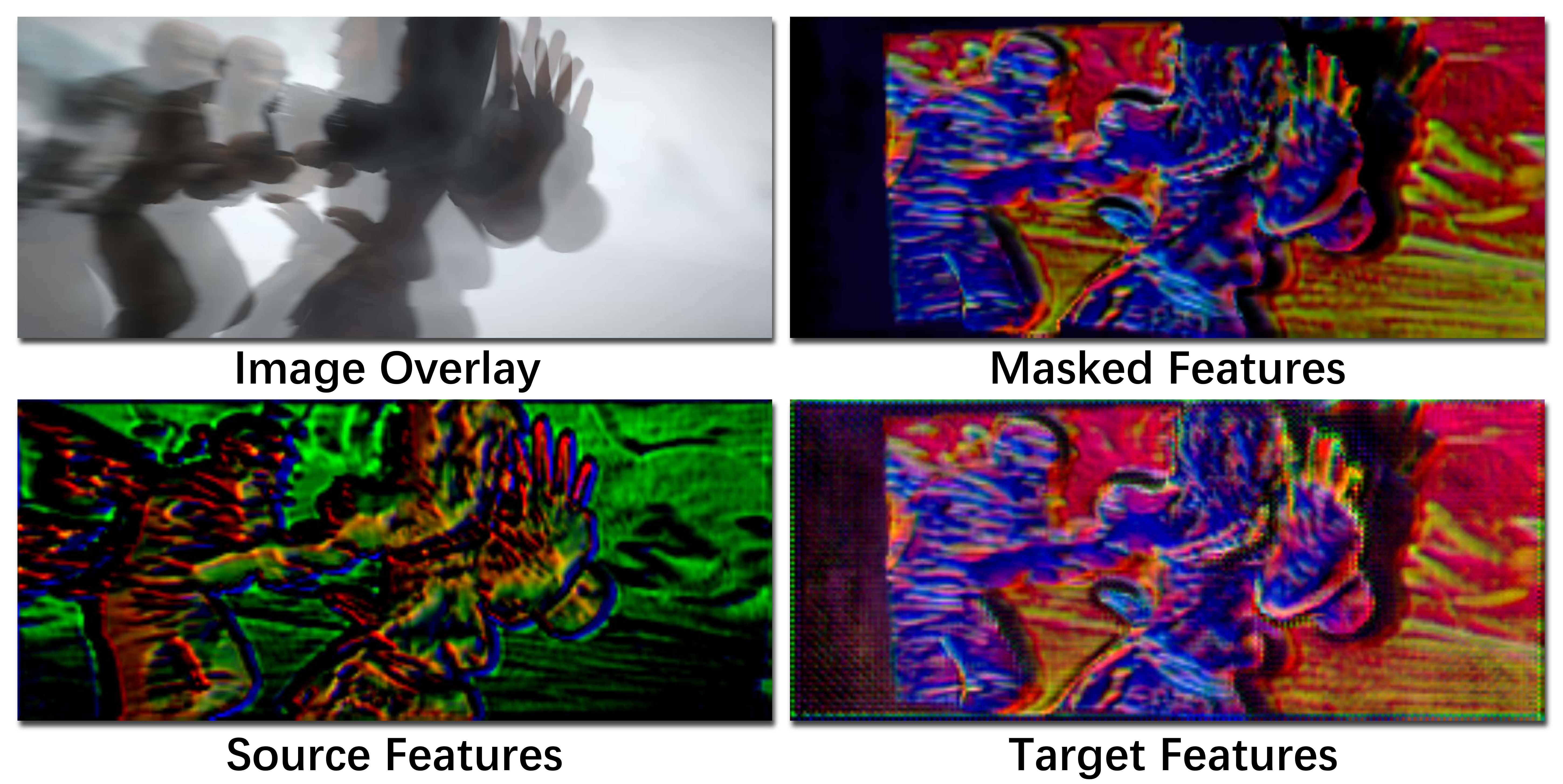}
\end{center}
   \caption{\textbf{Asymmetricity in the learned feature maps.} The source features and the target features are level-2 feature maps prior to the correlation layer in the AsymOFMM. This figure presents the input image overlay, image 1 features (source features), the warped image 2 features after masking (masked features) and after trade-off (target features). Comparing the source features with the target features, we can see that the AsymOFMM enables the network to learn very different feature representations.}
\label{fig:asym_features}
\end{figure}

\begin{figure}[t]
\begin{center}
   \includegraphics[width=1.0\linewidth]{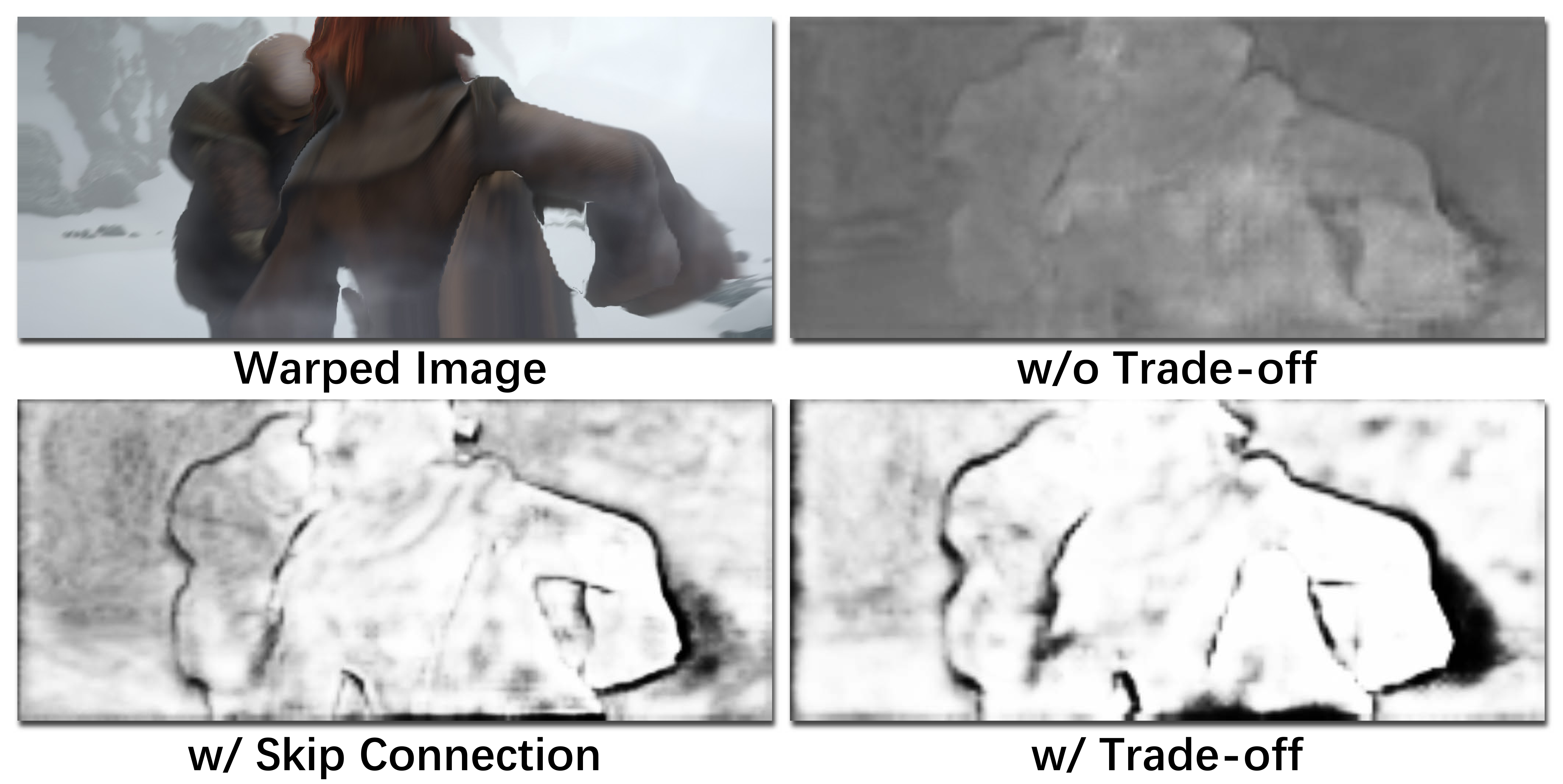}
\end{center}
   \caption{\textbf{The trade-off term facilitates the learning of occlusions.} Without the trade-off term, the learnable occlusion mask fails to achieve a clear estimation; if there is an additive shortcut that skips over warping, only motion boundaries are learned. With the trade-off term, large occlusions are successfully learned.}
\label{fig:mask_tradeoff}
\end{figure}

\subsection{Main Results}

MaskFlownet outperforms all published optical flow methods on the MPI Sintel~\cite{Butler2012Sintel}, KITTI 2012~\cite{Geiger2012KITTI} and KITTI 2015~\cite{Menze2015KITTI} benchmarks as presented in Table~\ref{tab:main}, while the end-to-end MaskFlownet-S achieves a satisfactory result as well. Values listed in the training sets only consider the models that have never seen them and hence comparable; note that training on the synthetic datasets is of limited generalizability.
Fig.~\ref{fig:flow_visualization} visualizes the predicted flows as a comparison. All samples are chosen from the Sintel training set (final pass). We can observe that MaskFlownet in general better separates moving objects from the background, and cascading significantly weakens the checkerboard effect while preserving clear object boundaries. Fig.~\ref{fig:mask_visualization} qualitatively demonstrates that the learnable occlusion mask matches the inverse ground-truth occlusion map fairly well, even if it is learned without any explicit supervision.

\begin{table}[t]
\scriptsize
\begin{center}
\begin{tabular}{lccccccc}
\toprule
\multirow{3}{*}{Module} & \multicolumn{3}{c}{Trained on Chairs} & \multicolumn{2}{c}{Things3D} & \multicolumn{2}{c}{Sintel} \\
\cmidrule(lr){2-4}
\cmidrule(lr){5-6}
\cmidrule(lr){7-8}
{} & Chairs & \multicolumn{2}{c}{Sintel (\emph{train})} & \multicolumn{2}{c}{Sintel (\emph{train})} & \multicolumn{2}{c}{Sintel (\emph{val})} \\
{} & \emph{test} & \emph{clean} & \emph{final} & \emph{clean} & \emph{final} & \emph{clean} & \emph{final} \\
\midrule
FMM & 1.61 & 3.25 & 4.59 & 2.55 & 4.05 & 3.02 & 4.70 \\
OFMM & 1.62 & 3.20 & 4.50 & 2.52 & 4.01 & 3.06 & 4.52 \\
AsymOFMM & \textbf{1.56} & \textbf{2.88} & \textbf{4.25} & \textbf{2.33} & \textbf{3.72} & \textbf{2.70} & \textbf{4.07} \\
\bottomrule
\end{tabular}
\end{center}
\caption{\textbf{Feature matching module.}}
\label{tab:FMM}
\end{table}

\begin{table}[t]
\scriptsize
\begin{center}
\begin{tabular}{lccccc}
\toprule
\multirow{3}{*}{Module} & \multicolumn{3}{c}{Trained on Chairs} & \multicolumn{2}{c}{Things3D} \\
\cmidrule(lr){2-4}
\cmidrule(lr){5-6}
{} & Chairs & \multicolumn{2}{c}{Sintel (\emph{train})} & \multicolumn{2}{c}{Sintel (\emph{train})} \\
{} & \emph{test} & \emph{clean} & \emph{final} & \emph{clean} & \emph{final} \\
\midrule
OFMM & 1.62 & 3.20 & 4.50 & 2.52 & 4.01 \\
+ sym-conv & 1.61 & 3.33 & 4.64 & 2.54 & 3.84 \\
+ asym-conv & \textbf{1.52} & 2.96 & 4.29 & 2.41 & 3.85 \\
+ deform-conv & 1.56 & \textbf{2.88} & \textbf{4.25} & \textbf{2.33} & \textbf{3.72} \\
\bottomrule
\end{tabular}
\end{center}
\caption{\textbf{Asymmetricity and deformable convolution.}}
\label{tab:asym}
\end{table}

\begin{table}[t]
\scriptsize
\begin{center}
\begin{tabular}{lccc}
\toprule
\multirow{3}{*}{Module (AsymOFMM)} & \multicolumn{3}{c}{Trained on Chairs} \\
\cmidrule(lr){2-4}
{} & Chairs & \multicolumn{2}{c}{Sintel (\emph{train})} \\
{} & \emph{test} & \emph{clean} & \emph{final} \\
\midrule
w/o mask w/o trade-off & 1.58 & 3.08 & 4.29 \\
w/ mask w/o trade-off & 1.60 & 3.06 & 4.32 \\
w/o mask w/ trade-off & 1.58 & 2.97 & 4.30 \\
(w/ mask w/ trade-off) & \textbf{1.56} & \textbf{2.88} & \textbf{4.25} \\
\bottomrule
\end{tabular}
\end{center}
\caption{\textbf{Learnable occlusion mask and the trade-off term.}}
\label{tab:mask}
\end{table}

\begin{table}[t]
\scriptsize
\begin{center}
\begin{tabular}{lcc}
\toprule
\multirow{3}{*}{Network} & \multicolumn{2}{c}{Tuned on Sintel} \\
\cmidrule(lr){2-3}
{} & \multicolumn{2}{c}{Sintel (\emph{val})} \\
{} & \emph{clean} & \emph{final} \\
\midrule
MaskFlownet-S & 2.70 & 4.07 \\
+ single pyramid w/o mask & 2.53 & 3.90 \\
+ single pyramid w/ mask & 2.55 & 3.88 \\
+ dual pyramids w/o mask & \textbf{2.52} & 3.85 \\
+ dual pyramids w/ mask & \textbf{2.52} & \textbf{3.83} \\
\bottomrule
\end{tabular}
\end{center}
\caption{\textbf{Network cascading with dual pyramids.}}
\label{tab:cascade}
\end{table}

\subsection{Ablation Study} \label{sec:ablation_asym}


\paragraph{Feature Matching Module.}

Table~\ref{tab:FMM} presents the results when replacing the AsymOFMM in MaskFlownet-S with OFMM or FMM. We split about 20\% sequences for validation\footnote{ambush\_2, ambush\_6, bamboo\_2, cave\_4, market\_6, temple\_2.} when fine-tuning on the Sintel training set. While OFMM achieves relative gains compared with the original FMM, the proposed AsymOFMM significantly outperforms the symmetric variants.


\paragraph{Asymmetricity.}

We demonstrate the effectiveness of the asymmetricity by comparing AsymOFMM (i.e., ``OFMM + deform-conv'', MaskFlownet-S) with the raw OFMM, ``OFMM + sym-conv'' (adding an extra convolutional layer at each feature pyramid level), and ``OFMM + asym-conv'' (adding an asymmetric convolutional layer prior to the warping operation at each level). As shown in Table~\ref{tab:asym}, increasing the depth of convolutional layers has a limited impact in the symmetric setting, while a simple asymmetric design achieves consistently better performance. It also indicates that our deformable convolution can be a better choice over the ``asym-conv'' version.
Although it is commonly believed that matched patterns should be embedded into similar feature vectors, we suggest that the network can really benefit from learning very different feature representations as visualized in Fig.~\ref{fig:asym_features}.

\begin{figure}[t]
\begin{center}
   \includegraphics[width=1.0\linewidth]{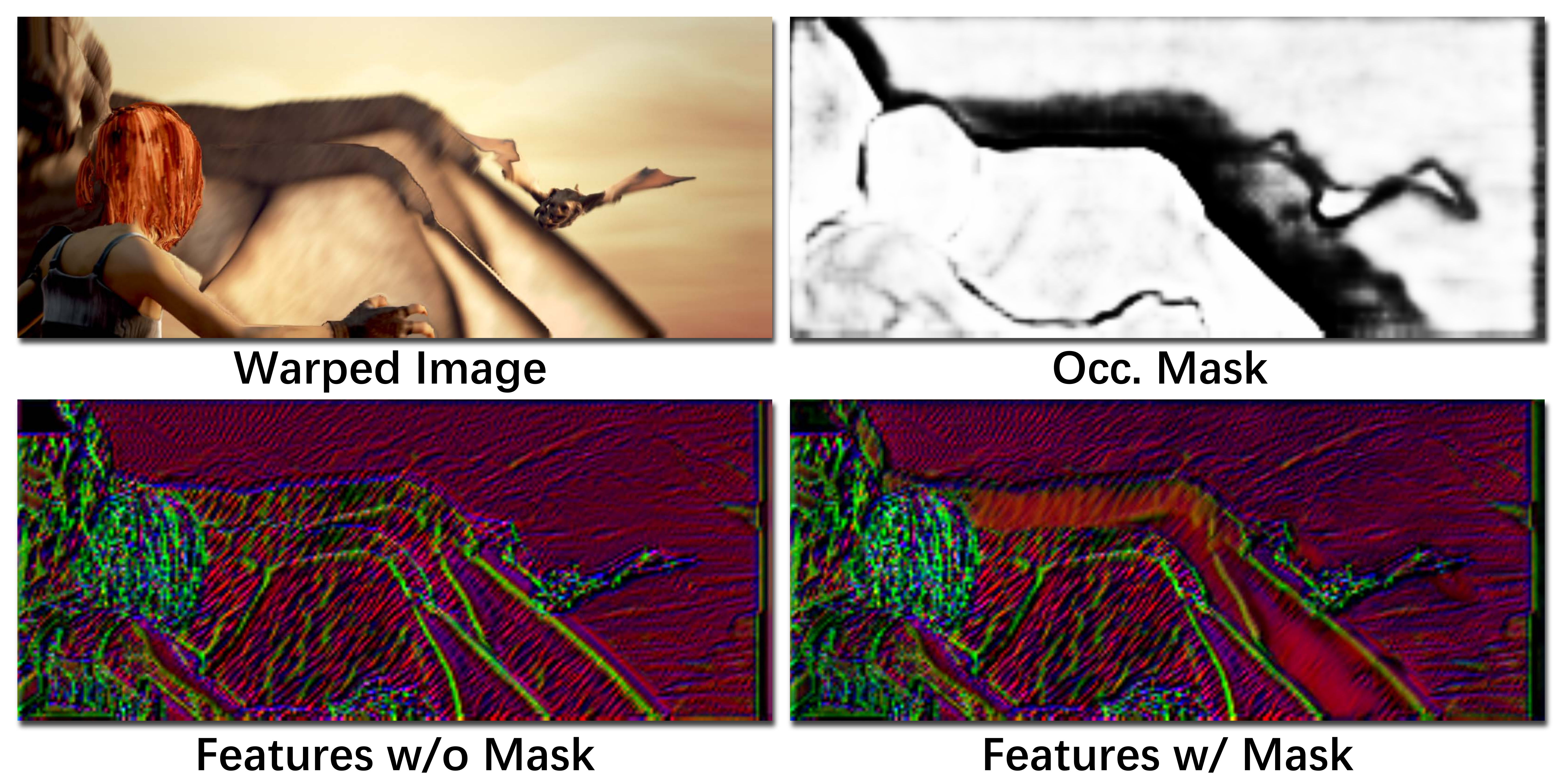}
\end{center}
   \caption{\textbf{Occlusion mask facilitates feature extraction in the occlusion-aware feature pyramid.} Comparing the learned features with and without (i.e., replaced by constant) mask, we can see that the occlusion mask draws a significant effect on smoothing the feature map at the occluded areas.}
\label{fig:occ_features}
\end{figure}

\paragraph{Learnable Occlusion Mask.}

Table~\ref{tab:mask} presents the results if the mask or the trade-off term is disabled. Interestingly, only the two factors combined lead to performance gains. A possible explanation is that the occlusion mask helps if and only if it is learned properly, where the trade-off term plays a vital role (see Fig.~\ref{fig:mask_tradeoff}).


\paragraph{Network Cascading.}

Table~\ref{tab:cascade} indicates that MaskFlownet consistently benefits from dual feature pyramids over a single new pyramid, while the concatenated occlusion mask gains the performance on the Sintel final pass. We hypothesize that the occlusion-aware feature pyramid mainly contributes to the harder final pass since the occluded areas can be more easily mismatched, but it might be overfitted on the easier clean pass. We demonstrate how the learned occlusion mask could affect the extracted feature map in Fig.~\ref{fig:occ_features}. The occluded areas are smoothened during feature extraction and hence become more distinguishable.

\section{Conclusion}

We propose the AsymOFMM, which incorporates a learnable occlusion mask that filters occluded areas immediately after feature warping without any explicit supervision. AsymOFMM can be easily integrated into an end-to-end network while introducing negligible computational cost. We further propose a two-stage network --- MaskFlownet --- which exploits dual pyramids and achieves superior performance on all modern optical flow benchmarks. Our approach opens a promisingly new perspective on dealing with occlusions for both supervised and unsupervised optical flow estimation, and we also expect it as an initiative and effective component in many other applications.


\clearpage

{\small
\bibliographystyle{ieee_fullname}
\bibliography{ref}

\begin{thebibliography}{10}\itemsep=-1pt

\bibitem{Bailer2017CNNMatch}
Christian Bailer, Kiran Varanasi, and Didier Stricker.
\newblock Cnn-based patch matching for optical flow with thresholded hinge
  embedding loss.
\newblock In {\em Proceedings of the IEEE Conference on Computer Vision and
  Pattern Recognition}, pages 3250--3259, 2017.

\bibitem{Bonneel2015Video}
Nicolas Bonneel, James Tompkin, Kalyan Sunkavalli, Deqing Sun, Sylvain Paris,
  and Hanspeter Pfister.
\newblock Blind video temporal consistency.
\newblock {\em ACM Transactions on Graphics (TOG)}, 34(6):196, 2015.

\bibitem{Brox2004High}
Thomas Brox, Andr{\'e}s Bruhn, Nils Papenberg, and Joachim Weickert.
\newblock High accuracy optical flow estimation based on a theory for warping.
\newblock In {\em European conference on computer vision}, pages 25--36.
  Springer, 2004.

\bibitem{Brox2010DescripMatch}
Thomas Brox and Jitendra Malik.
\newblock Large displacement optical flow: descriptor matching in variational
  motion estimation.
\newblock {\em IEEE transactions on pattern analysis and machine intelligence},
  33(3):500--513, 2010.

\bibitem{Butler2012Sintel}
Daniel~J Butler, Jonas Wulff, Garrett~B Stanley, and Michael~J Black.
\newblock A naturalistic open source movie for optical flow evaluation.
\newblock In {\em European conference on computer vision}, pages 611--625.
  Springer, 2012.

\bibitem{Chen2015MXNet}
Tianqi Chen, Mu Li, Yutian Li, Min Lin, Naiyan Wang, Minjie Wang, Tianjun Xiao,
  Bing Xu, Chiyuan Zhang, and Zheng Zhang.
\newblock Mxnet: A flexible and efficient machine learning library for
  heterogeneous distributed systems.
\newblock {\em arXiv preprint arXiv:1512.01274}, 2015.

\bibitem{Dai2017Deformable}
Jifeng Dai, Haozhi Qi, Yuwen Xiong, Yi Li, Guodong Zhang, Han Hu, and Yichen
  Wei.
\newblock Deformable convolutional networks.
\newblock In {\em Proceedings of the IEEE international conference on computer
  vision}, pages 764--773, 2017.

\bibitem{Dosovitskiy2015Flownet}
Alexey Dosovitskiy, Philipp Fischer, Eddy Ilg, Philip Hausser, Caner Hazirbas,
  Vladimir Golkov, Patrick Van Der~Smagt, Daniel Cremers, and Thomas Brox.
\newblock Flownet: Learning optical flow with convolutional networks.
\newblock In {\em Proceedings of the IEEE international conference on computer
  vision}, pages 2758--2766, 2015.

\bibitem{Geiger2012KITTI}
Andreas Geiger, Philip Lenz, and Raquel Urtasun.
\newblock Are we ready for autonomous driving? the kitti vision benchmark
  suite.
\newblock In {\em 2012 IEEE Conference on Computer Vision and Pattern
  Recognition}, pages 3354--3361. IEEE, 2012.

\bibitem{Horn1981Flow}
Berthold~KP Horn and Brian~G Schunck.
\newblock Determining optical flow.
\newblock {\em Artificial intelligence}, 17(1-3):185--203, 1981.

\bibitem{Hui2018Liteflownet}
Tak-Wai Hui, Xiaoou Tang, and Chen Change~Loy.
\newblock Liteflownet: A lightweight convolutional neural network for optical
  flow estimation.
\newblock In {\em Proceedings of the IEEE Conference on Computer Vision and
  Pattern Recognition}, pages 8981--8989, 2018.

\bibitem{Hui2019Liteflownet2}
Tak-Wai Hui, Xiaoou Tang, and Chen~Change Loy.
\newblock A lightweight optical flow cnn--revisiting data fidelity and
  regularization.
\newblock {\em arXiv preprint arXiv:1903.07414}, 2019.

\bibitem{Hur2019IRR}
Junhwa Hur and Stefan Roth.
\newblock Iterative residual refinement for joint optical flow and occlusion
  estimation.
\newblock In {\em Proceedings of the IEEE Conference on Computer Vision and
  Pattern Recognition}, pages 5754--5763, 2019.

\bibitem{Ilg2017Flownet2}
Eddy Ilg, Nikolaus Mayer, Tonmoy Saikia, Margret Keuper, Alexey Dosovitskiy,
  and Thomas Brox.
\newblock Flownet 2.0: Evolution of optical flow estimation with deep networks.
\newblock In {\em Proceedings of the IEEE conference on computer vision and
  pattern recognition}, pages 2462--2470, 2017.

\bibitem{Ilg2018DispNet2}
Eddy Ilg, Tonmoy Saikia, Margret Keuper, and Thomas Brox.
\newblock Occlusions, motion and depth boundaries with a generic network for
  disparity, optical flow or scene flow estimation.
\newblock In {\em Proceedings of the European Conference on Computer Vision
  (ECCV)}, pages 614--630, 2018.

\bibitem{Janai2018UnMulti}
Joel Janai, Fatma Guney, Anurag Ranjan, Michael Black, and Andreas Geiger.
\newblock Unsupervised learning of multi-frame optical flow with occlusions.
\newblock In {\em Proceedings of the European Conference on Computer Vision
  (ECCV)}, pages 690--706, 2018.

\bibitem{Jason2016UnBrightness}
J~Yu Jason, Adam~W Harley, and Konstantinos~G Derpanis.
\newblock Back to basics: Unsupervised learning of optical flow via brightness
  constancy and motion smoothness.
\newblock In {\em European Conference on Computer Vision}, pages 3--10.
  Springer, 2016.

\bibitem{Liu2019SelFlow}
Pengpeng Liu, Michael Lyu, Irwin King, and Jia Xu.
\newblock Selflow: Self-supervised learning of optical flow.
\newblock In {\em Proceedings of the IEEE Conference on Computer Vision and
  Pattern Recognition}, pages 4571--4580, 2019.

\bibitem{Mayer2016Things3D}
Nikolaus Mayer, Eddy Ilg, Philip Hausser, Philipp Fischer, Daniel Cremers,
  Alexey Dosovitskiy, and Thomas Brox.
\newblock A large dataset to train convolutional networks for disparity,
  optical flow, and scene flow estimation.
\newblock In {\em Proceedings of the IEEE Conference on Computer Vision and
  Pattern Recognition}, pages 4040--4048, 2016.

\bibitem{Meister2018UnFlow}
Simon Meister, Junhwa Hur, and Stefan Roth.
\newblock Unflow: Unsupervised learning of optical flow with a bidirectional
  census loss.
\newblock In {\em Thirty-Second AAAI Conference on Artificial Intelligence},
  2018.

\bibitem{Memin1998Dense}
Etienne M{\'e}min and Patrick P{\'e}rez.
\newblock Dense estimation and object-based segmentation of the optical flow
  with robust techniques.
\newblock {\em IEEE Transactions on Image Processing}, 7(5):703--719, 1998.

\bibitem{Menze2015Driving}
Moritz Menze and Andreas Geiger.
\newblock Object scene flow for autonomous vehicles.
\newblock In {\em Proceedings of the IEEE Conference on Computer Vision and
  Pattern Recognition}, pages 3061--3070, 2015.

\bibitem{Menze2015KITTI}
Moritz Menze and Andreas Geiger.
\newblock Object scene flow for autonomous vehicles.
\newblock In {\em Proceedings of the IEEE Conference on Computer Vision and
  Pattern Recognition}, pages 3061--3070, 2015.

\bibitem{Neoral2018Continual}
Michal Neoral, Jan {\v{S}}ochman, and Ji{\v{r}}{\'\i} Matas.
\newblock Continual occlusion and optical flow estimation.
\newblock In {\em Asian Conference on Computer Vision}, pages 159--174.
  Springer, 2018.

\bibitem{pytorch-pwc}
Simon Niklaus.
\newblock A reimplementation of {PWC-Net} using {PyTorch}.
\newblock \url{https://github.com/sniklaus/pytorch-pwc}, 2018.

\bibitem{Pang2019PedMaskAtt}
Yanwei Pang, Jin Xie, Muhammad~Haris Khan, Rao~Muhammad Anwer, Fahad~Shahbaz
  Khan, and Ling Shao.
\newblock Mask-guided attention network for occluded pedestrian detection.
\newblock In {\em Proceedings of the IEEE International Conference on Computer
  Vision}, pages 4967--4975, 2019.

\bibitem{Ranjan2017SpyNet}
Anurag Ranjan and Michael~J Black.
\newblock Optical flow estimation using a spatial pyramid network.
\newblock In {\em Proceedings of the IEEE Conference on Computer Vision and
  Pattern Recognition}, pages 4161--4170, 2017.

\bibitem{Ren2017UnDeep}
Zhe Ren, Junchi Yan, Bingbing Ni, Bin Liu, Xiaokang Yang, and Hongyuan Zha.
\newblock Unsupervised deep learning for optical flow estimation.
\newblock In {\em Thirty-First AAAI Conference on Artificial Intelligence},
  2017.

\bibitem{Simonyan2014Action}
Karen Simonyan and Andrew Zisserman.
\newblock Two-stream convolutional networks for action recognition in videos.
\newblock In {\em Advances in neural information processing systems}, pages
  568--576, 2014.

\bibitem{Song2019FaceOcc}
Lingxue Song, Dihong Gong, Zhifeng Li, Changsong Liu, and Wei Liu.
\newblock Occlusion robust face recognition based on mask learning with
  pairwise differential siamese network.
\newblock In {\em Proceedings of the IEEE International Conference on Computer
  Vision}, pages 773--782, 2019.

\bibitem{Sun2018PWC+}
Deqing Sun, Xiaodong Yang, Ming-Yu Liu, and Jan Kautz.
\newblock Models matter, so does training: An empirical study of cnns for
  optical flow estimation.
\newblock {\em arXiv preprint arXiv:1809.05571}, 2018.

\bibitem{Sun2018PWC}
Deqing Sun, Xiaodong Yang, Ming-Yu Liu, and Jan Kautz.
\newblock Pwc-net: Cnns for optical flow using pyramid, warping, and cost
  volume.
\newblock In {\em Proceedings of the IEEE Conference on Computer Vision and
  Pattern Recognition}, pages 8934--8943, 2018.

\bibitem{Sundaram2010Dense}
Narayanan Sundaram, Thomas Brox, and Kurt Keutzer.
\newblock Dense point trajectories by gpu-accelerated large displacement
  optical flow.
\newblock In {\em European conference on computer vision}, pages 438--451.
  Springer, 2010.

\bibitem{Wang2018NonLocal}
Xiaolong Wang, Ross Girshick, Abhinav Gupta, and Kaiming He.
\newblock Non-local neural networks.
\newblock In {\em Proceedings of the IEEE Conference on Computer Vision and
  Pattern Recognition}, pages 7794--7803, 2018.

\bibitem{Wang2018Occ}
Yang Wang, Yi Yang, Zhenheng Yang, Liang Zhao, Peng Wang, and Wei Xu.
\newblock Occlusion aware unsupervised learning of optical flow.
\newblock In {\em Proceedings of the IEEE Conference on Computer Vision and
  Pattern Recognition}, pages 4884--4893, 2018.

\bibitem{Wedel2009Structure}
Andreas Wedel, Daniel Cremers, Thomas Pock, and Horst Bischof.
\newblock Structure-and motion-adaptive regularization for high accuracy optic
  flow.
\newblock In {\em 2009 IEEE 12th International Conference on Computer Vision},
  pages 1663--1668. IEEE, 2009.

\bibitem{Weinzaepfel2013DeepFlow}
Philippe Weinzaepfel, Jerome Revaud, Zaid Harchaoui, and Cordelia Schmid.
\newblock Deepflow: Large displacement optical flow with deep matching.
\newblock In {\em Proceedings of the IEEE International Conference on Computer
  Vision}, pages 1385--1392, 2013.

\bibitem{Wulff2017MRFlow}
Jonas Wulff, Laura Sevilla-Lara, and Michael~J Black.
\newblock Optical flow in mostly rigid scenes.
\newblock In {\em Proceedings of the IEEE Conference on Computer Vision and
  Pattern Recognition}, pages 4671--4680, 2017.

\bibitem{Xu2017DirectCost}
Jia Xu, Ren{\'e} Ranftl, and Vladlen Koltun.
\newblock Accurate optical flow via direct cost volume processing.
\newblock In {\em Proceedings of the IEEE Conference on Computer Vision and
  Pattern Recognition}, pages 1289--1297, 2017.

\bibitem{Yang2019VCN}
Gengshan Yang and Deva Ramanan.
\newblock Volumetric correspondence networks for optical flow.
\newblock In {\em Advances in neural information processing systems}, 2019.

\bibitem{Zhang2018OccRCNN}
Shifeng Zhang, Longyin Wen, Xiao Bian, Zhen Lei, and Stan~Z Li.
\newblock Occlusion-aware r-cnn: detecting pedestrians in a crowd.
\newblock In {\em Proceedings of the European Conference on Computer Vision
  (ECCV)}, pages 637--653, 2018.

\bibitem{Zhang2018PedGuidedAtt}
Shanshan Zhang, Jian Yang, and Bernt Schiele.
\newblock Occluded pedestrian detection through guided attention in cnns.
\newblock In {\em Proceedings of the IEEE Conference on Computer Vision and
  Pattern Recognition}, pages 6995--7003, 2018.

\end{thebibliography}
}

\clearpage

\begin{appendices}

\section{More Implementation Details}

\paragraph{Training Schedule.}

\begin{figure}[ht]
\begin{center}
\subfigure[Schedule for fine-tuning on Sintel.]{ \label{fig:lr_sintel}
   \includegraphics[width=1.0\linewidth]{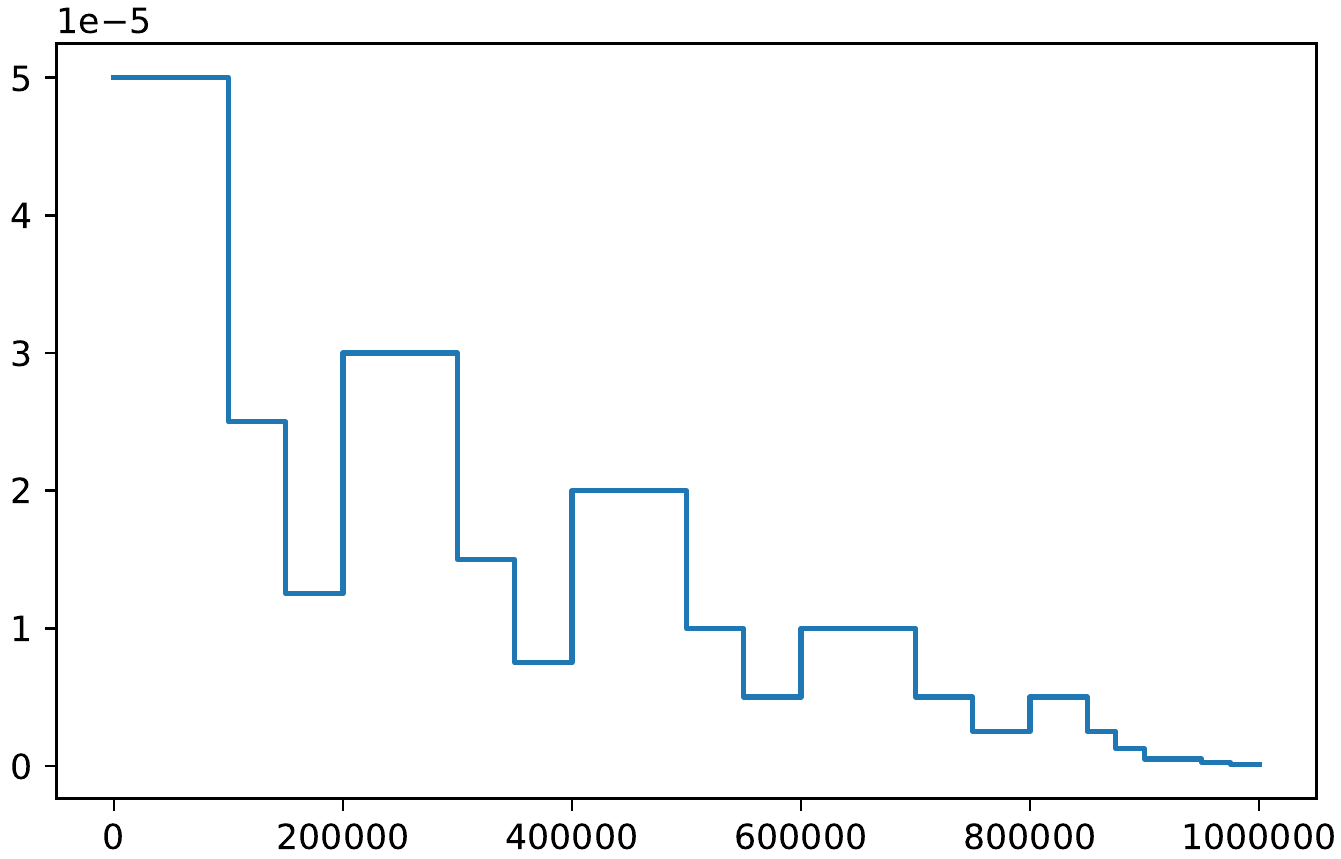}}
\subfigure[A shorter schedule for the second stage on FlyingChairs.]{ \label{fig:lr_chairs_short}
   \includegraphics[width=1.0\linewidth]{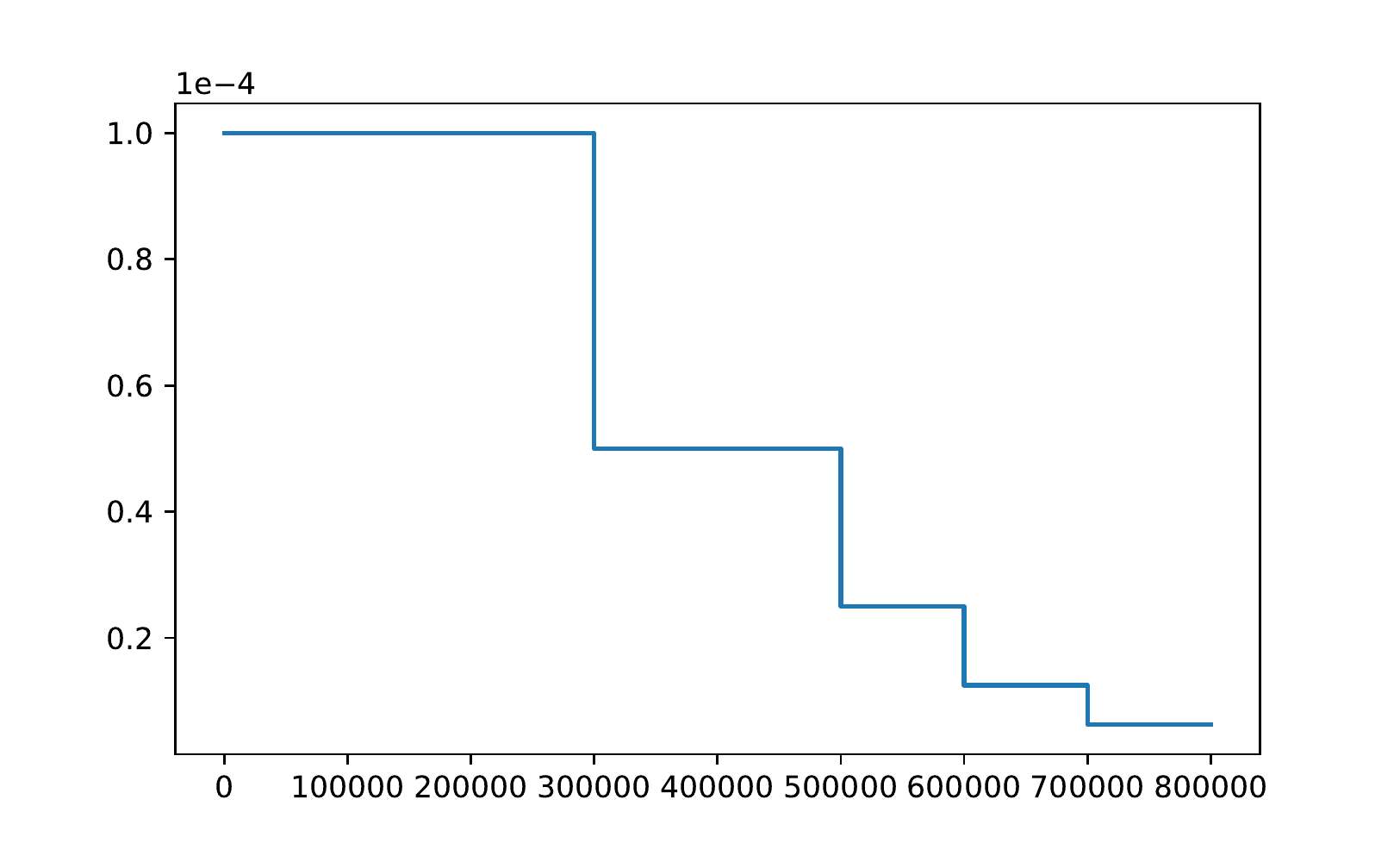}}
\end{center}
   \caption{\textbf{Learning rate schedules.}}
\label{fig:lr}
\end{figure}

When fine-tuning on Sintel, we use a longer schedule (see Fig.~\ref{fig:lr_sintel}) referring to the cyclic learning rate proposed by PWC-Net+~\cite{Sun2018PWC+}. When training the second stage, we follow again the same schedule as the first stage for all datasets except that it is shorter on FlyingChairs (see Fig.~\ref{fig:lr_chairs_short}). For submission to the test set, we train on the whole training set and reduce randomness by averaging 3 independent runs due to the huge variance.

\paragraph{Data Augmentation.}


We implement geometric and chromatic augmentations referring to the implementation of FlowNet~\cite{Dosovitskiy2015Flownet} and IRR-PWC~\cite{Hur2019IRR}. Details about the sampling ranges for each training stage are provided in Table~\ref{tab:geo_aug} (for geometric augmentations) and Table~\ref{tab:chr_aug} (for chromatic augmentations). We use the same augmentations on FlyingThings3D as FlyingChairs. We finally apply a random crop (within valid areas) using a size of $448 \times 320$ on FlyingChairs, $768 \times 384$ on FlyingThings3D, $768 \times 320$ on Sintel, and $896 \times 320$ on KITTI. To avoid out-of-bound areas after cropping, we compute the minimum degree of zoom that forces the existence of a valid crop.

\begin{table}[t]
\small
\begin{center}
\begin{tabular}{lccc}
\toprule
Geometric Aug.\@ & Chairs & Sintel & KITTI \\
\midrule
Horizontal Flip & 0.5 & 0.5 & 0.5 \\
Squeeze & 0.9 & 0.9 & 0.95 \\
Translation & 0.1 & 0.1 & 0.05 \\
Rel.\@ Translation & 0.025 & 0.025 & 0.0125 \\
Rotation & $\ang{17}$ & $\ang{17}$ & $\ang{5}$ \\
Rel.\@ Rotation & $\ang{4.25}$ & $\ang{4.25}$ & $\ang{1.25}$ \\
Zoom & $[0.9, 2.0]$ & $[0.9, 1.5]$ & $[0.95, 1.25]$ \\
Rel.\@ Zoom & 0.96 & 0.96 & 0.98 \\
\bottomrule
\end{tabular}
\end{center}
\caption{\textbf{Geometric augmentations.}}
\label{tab:geo_aug}
\end{table}

\begin{table}[t]
\small
\begin{center}
\begin{tabular}{lccc}
\toprule
Chromatic Aug.\@ & Chairs & Sintel & KITTI \\
\midrule
Contrast & $[-0.4, 0.8]$ & $[-0.4, 0.8]$ & $[-0.2, 0.4]$ \\
Brightness & 0.1 & 0.1 & 0.05 \\
Channel & $[0.8, 1.4]$ & $[0.8, 1.4]$ & $[0.9, 1.2]$ \\
Saturation & 0.5 & 0.5 & 0.25 \\
Hue & 0.5 & 0.5 & 0.1 \\
Noise & 0.04 & 0 & 0.02 \\
\bottomrule
\end{tabular}
\end{center}
\caption{\textbf{Chromatic augmentations.}}
\label{tab:chr_aug}
\end{table}

\section{More Visualizations}

\begin{figure*}[t]
\begin{center}
   \includegraphics[width=1.0\linewidth]{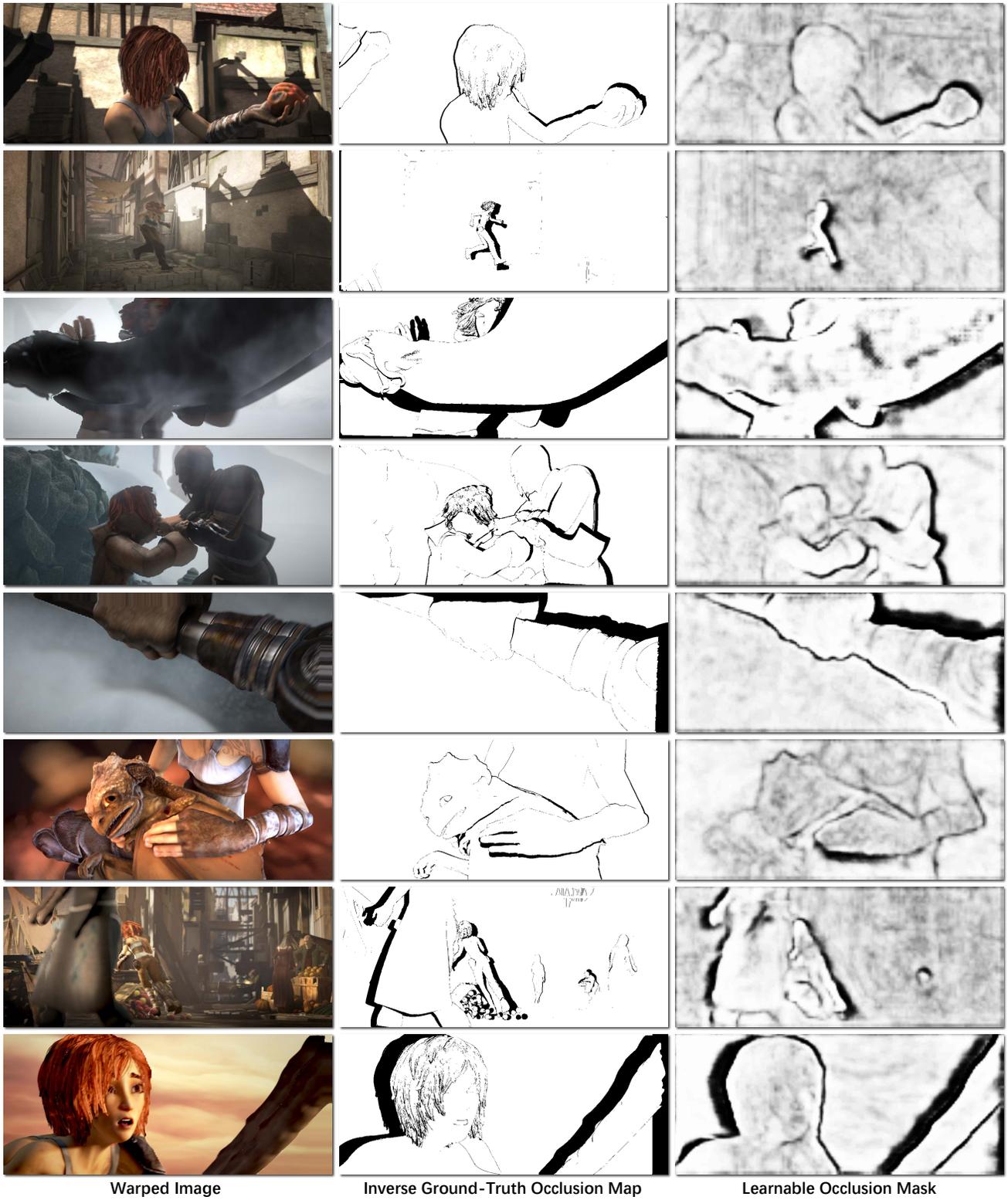}
\end{center}
   \caption{\textbf{More visualizations of the learnable occlusion mask.} All samples are chosen from the Sintel training set (final pass). The learnable occlusion masks are expected to (roughly) match the inverse ground-truth occlusion maps, even if they are learned without any explicit supervision.}
\label{fig:mask_visualization_suppl}
\end{figure*}

\begin{figure*}[t]
\begin{center}
   \includegraphics[width=1.0\linewidth]{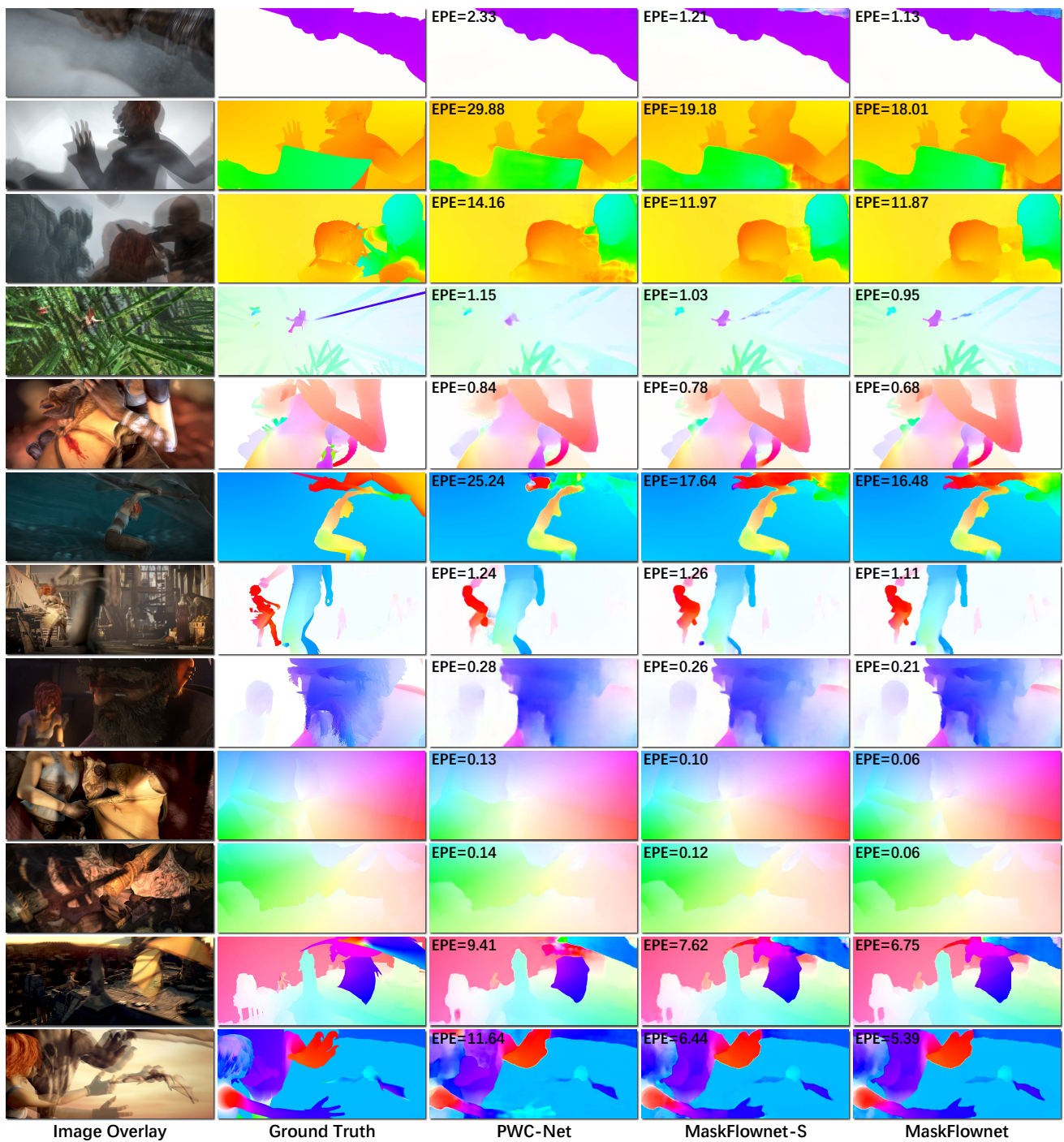}
\end{center}
   \caption{\textbf{More visualizations for qualitative comparison among PWC-Net~\cite{Sun2018PWC}, MaskFlownet-S, and MaskFlownet.} All samples are chosen from the Sintel training set (final pass). We replicate PWC-Net using the PyTorch reimplementation~\cite{pytorch-pwc} that provides a pretrained model of the ``PWC-Net\_ROB'' version~\cite{Sun2018PWC+}.}
\label{fig:flow_visualization_suppl}
\end{figure*}

More visualizations of the learnable occlusion mask and the flow predictions are presented in Fig.~\ref{fig:mask_visualization_suppl} and Fig.~\ref{fig:flow_visualization_suppl}. Note that the learned occlusion masks are relatively vague at the image boundary, since the network cannot learn to mask out-of-bound features that are already zeros. We expect that the estimation results can be further improved if out-of-bound areas are manually regarded as occlusions.

\section{Screenshots on Benchmarks}

\begin{figure*}[t]
\begin{center}
   \includegraphics[width=0.85\linewidth]{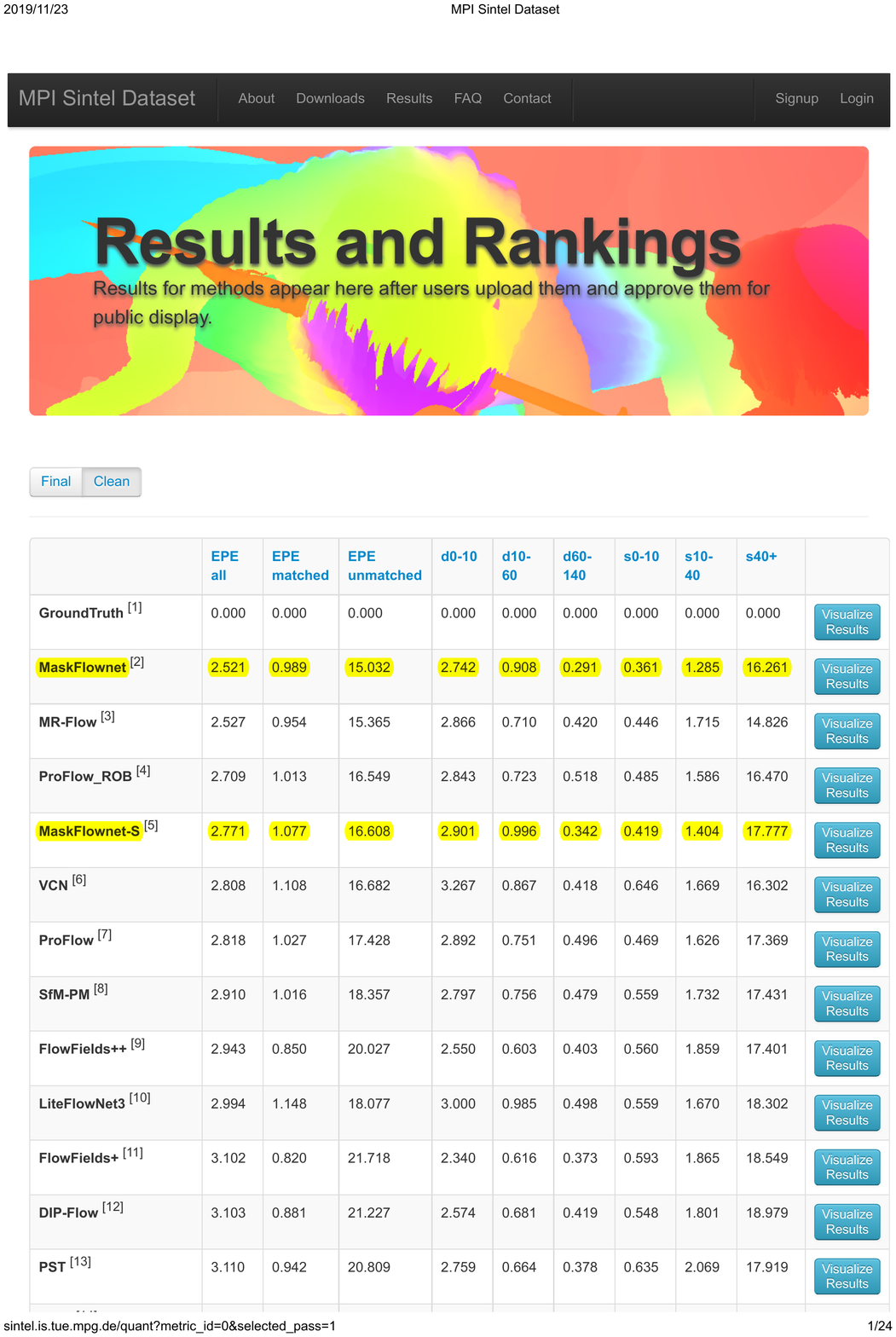}
\end{center}
   \caption{\textbf{Screenshot on the MPI Sintel clean pass (printed as PDF).}}
\label{fig:screenshot_sintel_clean}
\end{figure*}

\begin{figure*}[t]
\begin{center}
   \includegraphics[width=0.85\linewidth]{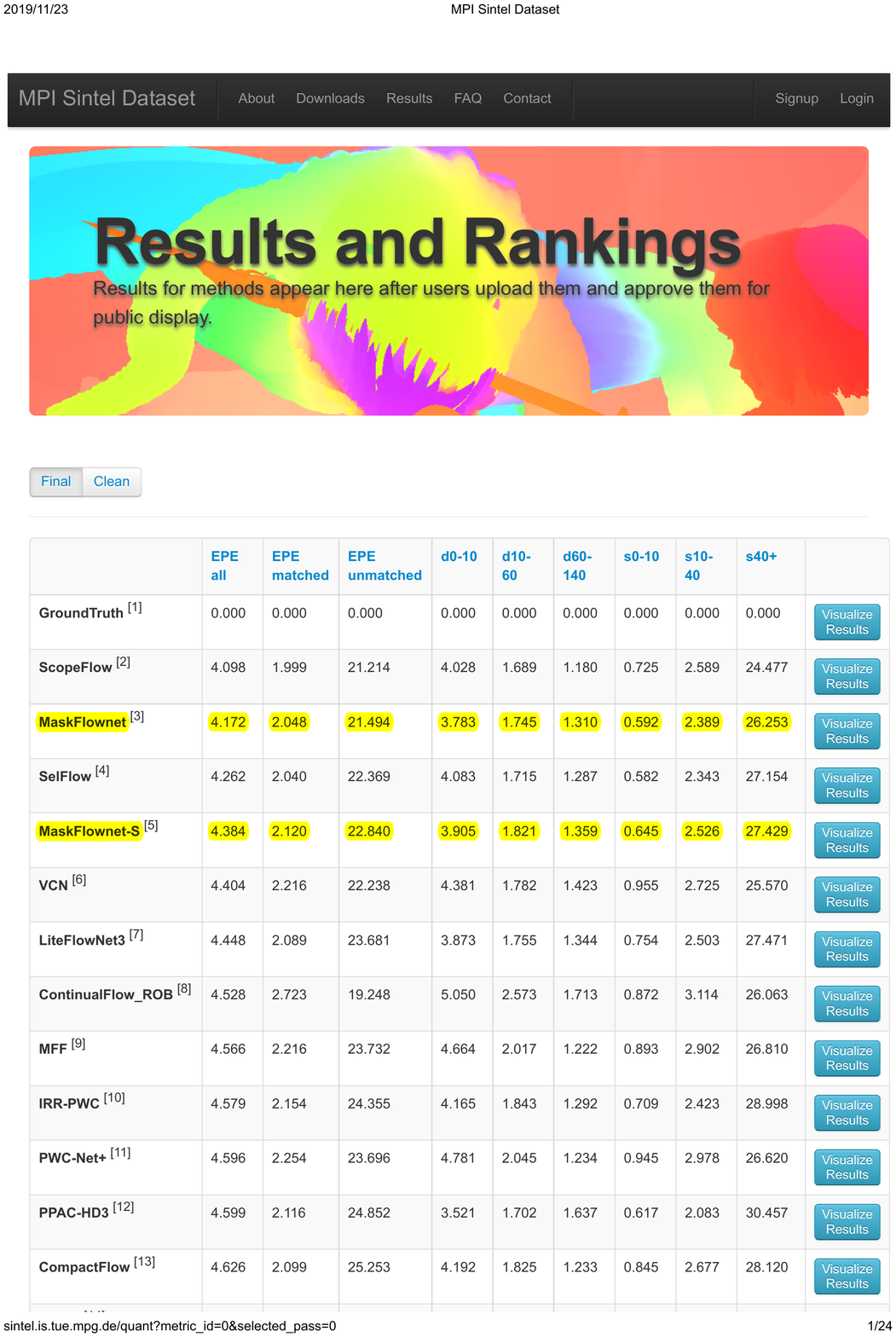}
\end{center}
   \caption{\textbf{Screenshot on MPI Sintel final pass (printed as PDF).}}
\label{fig:screenshot_sintel_final}
\end{figure*}

At the time of submission, MaskFlownet ranks first on the MPI Sintel benchmark on both clean pass (see Fig.~\ref{fig:screenshot_sintel_clean}) and final pass (see Fig.~\ref{fig:screenshot_sintel_final}). Note that the top entry (ScopeFlow) at the time of screenshot (Nov.~23th, 2019) on the final pass is a new anonymous submission, with a relatively poor performance on the clean pass. Remarkably, MaskFlownet outperforms the previous top entry on the clean pass (MR-Flow~\cite{Wulff2017MRFlow}) that uses the rigidity assumption while being very slow, as well as the previous top entry on the final pass (SelFlow~\cite{Liu2019SelFlow}) that uses multi-frame inputs.

\begin{figure*}[t]
\begin{center}
   \includegraphics[width=0.85\linewidth]{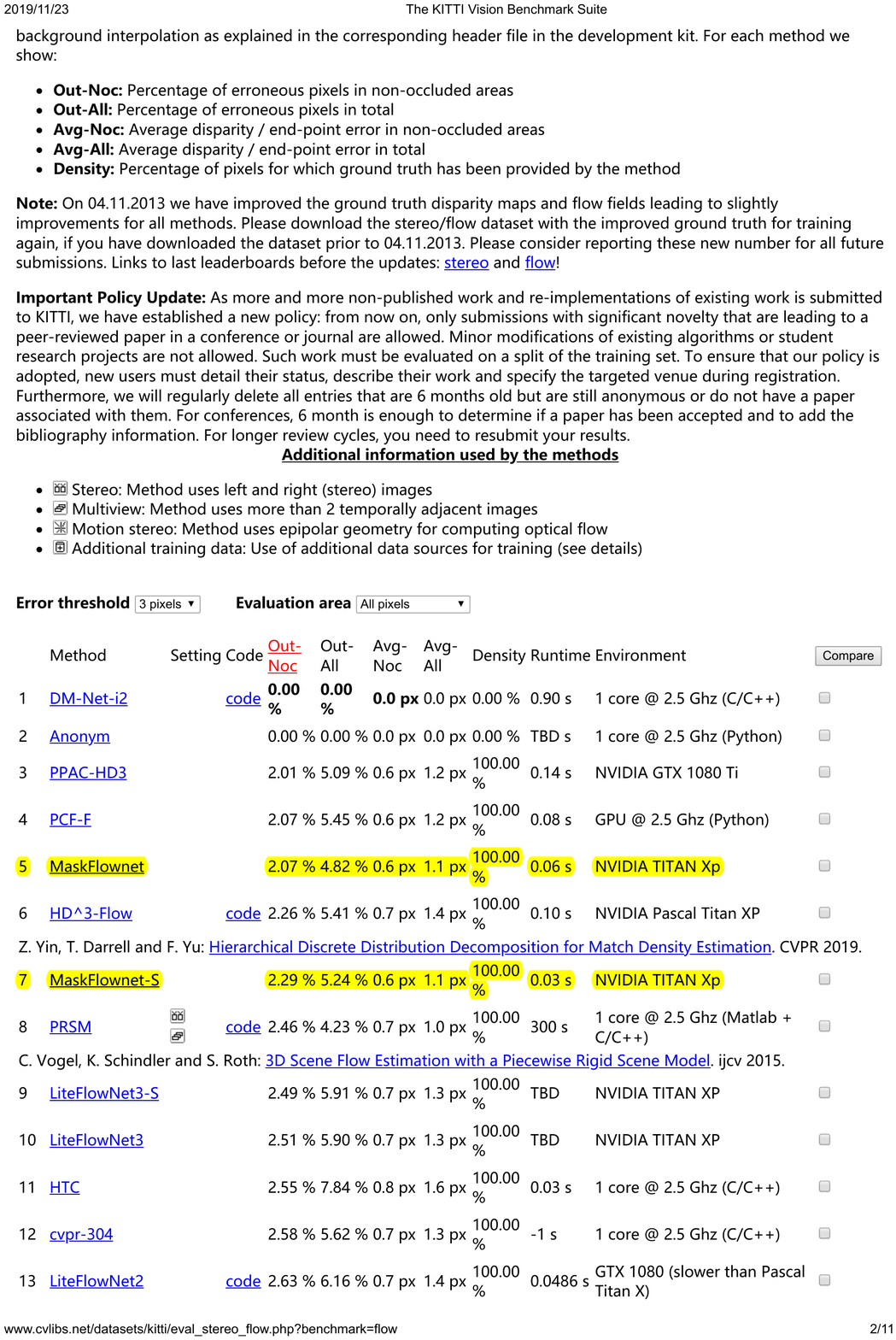}
\end{center}
   \caption{\textbf{Screenshot on the KITTI 2012 benchmark (printed as PDF).}}
\label{fig:screenshot_kitti_12}
\end{figure*}

\begin{figure*}[t]
\begin{center}
   \includegraphics[width=0.85\linewidth]{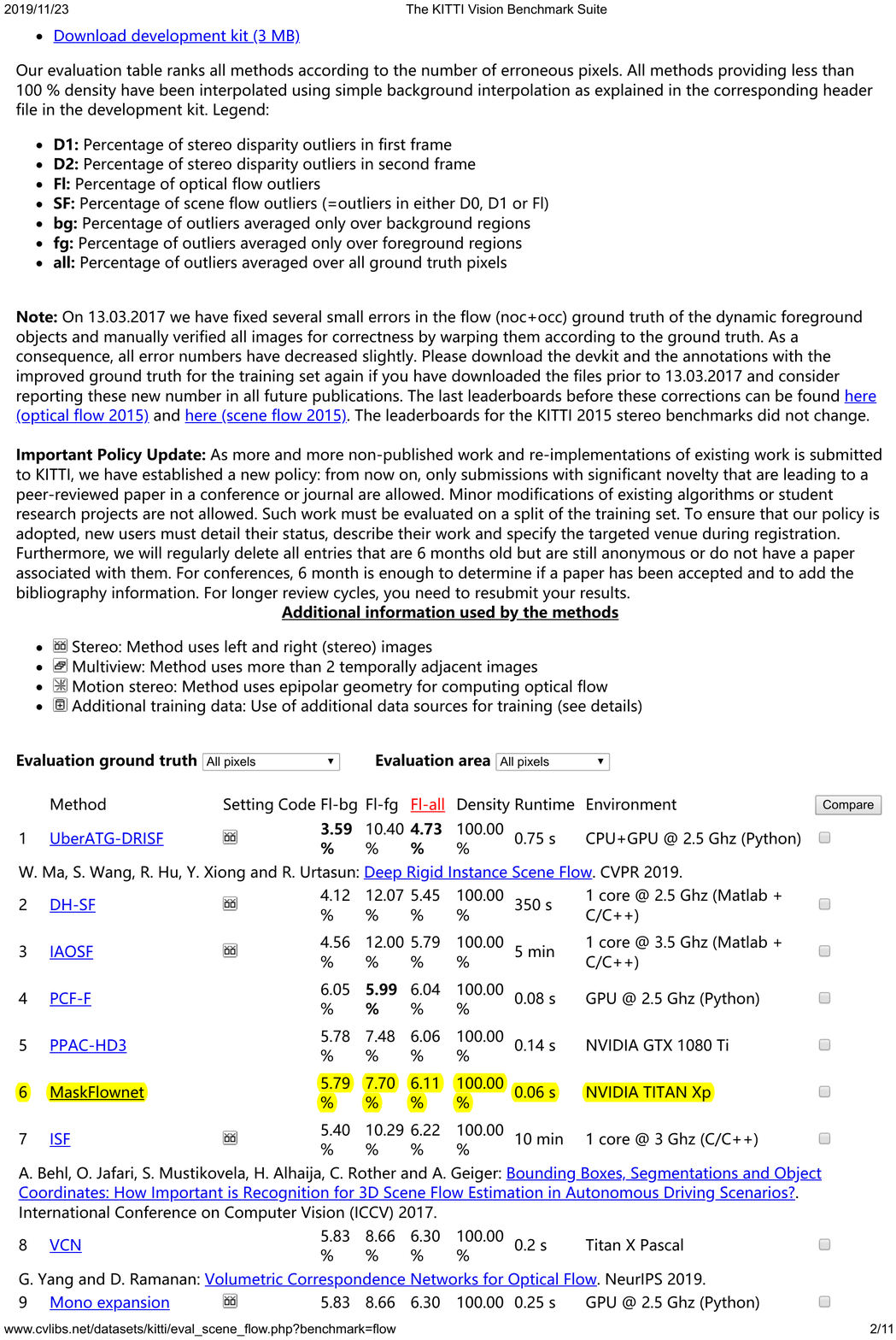}
\end{center}
   \caption{\textbf{Screenshot on the KITTI 2015 benchmark (printed as PDF).} MaskFlownet-S ranks 14th.}
\label{fig:screenshot_kitti_15}
\end{figure*}

On the KITTI 2012 and 2015 benchmarks, MaskFlownet surpasses all optical flow methods (excluding the anonymous entries) at the time of submission (see Fig.~\ref{fig:screenshot_kitti_12} and Fig.~\ref{fig:screenshot_kitti_15}). Note that the top 3 entries on the KITTI 2015 benchmark are \emph{scene flow} methods that use stereo images and thus not comparable.

\end{appendices}

\end{document}